\documentclass[10pt]{article}

\usepackage[utf8]{inputenc}
\usepackage[a4paper,lmargin=1.5in, rmargin=1.5in, footnotesep=1cm plus 4pt minus 4pt]{geometry}
\usepackage{setspace}
\usepackage{amsmath,amsthm,amssymb,esint}
\usepackage{mathtools, nccmath}
\allowdisplaybreaks
\usepackage{multirow}
\usepackage{sansmath}
\usepackage{pgfplots}
\pgfplotsset{compat = newest}
\usepackage{listings}
\usepackage{color}
\usepackage{placeins}
\usepackage{multicol}
\usepackage{multirow}
\usepackage{enumitem}
\usepackage{lipsum}
\usepackage{cases}
\usepackage{epstopdf}
\usepackage{float}
\usepackage{booktabs}
\usepackage{dirtytalk}
\usepackage{subcaption}
\usepackage{graphicx}

\usepackage{comment}

\usepackage[T1]{fontenc}
\usepackage{lmodern}
\usepackage{times}

\usepackage{sectsty}
\allsectionsfont{\sffamily\bfseries}
\subsubsectionfont{\sffamily\bfseries}

\usepackage[font=sf,labelfont=bf]{caption}

\usepackage[font=sf]{caption}

\newtheoremstyle{mystyle1}
  {}
  {}
  {\itshape}
  {}
  {\sffamily\bfseries}
  {{\sffamily\mdseries .}}
  { }
  {}
\theoremstyle{mystyle1}

\newtheorem{lemm}{Lemma}
\newtheorem{prop}{Proposition}

\newtheorem{examp}{Example}

\newtheoremstyle{mystyle2}
  {}
  {}
  {}
  {}
  {\itshape\sffamily}
  {\textsf{.}}
  { }
  {}
\theoremstyle{mystyle2}

\newtheorem*{rem}{Remark}

\usepackage{hyperref}
\usepackage{xcolor}
\definecolor{rsrs}{RGB}{19, 59, 123}
\definecolor{myred}{rgb}{0.7, 0.0, 0.0}
\definecolor{mygreen}{rgb}{0.0, 0.3, 0.0}
\hypersetup{
    colorlinks,
    linkcolor={rsrs},
    citecolor={rsrs},
    urlcolor={rsrs}
}

\usepackage{tocloft}

\usepackage[nottoc]{tocbibind}

\usepackage{kantlipsum}

\usepackage[
    backend=biber,
    citestyle=authoryear-icomp,
    bibstyle=authoryear,
    natbib=true,
    maxbibnames=99,
    maxcitenames=2,
    giveninits=true,
    terseinits=true,
    uniquelist=false
]{biblatex}
\DefineBibliographyStrings{english}{%
    andothers = {\em et\addabbrvspace al.}
}
\DeclareNameAlias{author}{last-first}
\DeclareFieldFormat*{title}{#1}
\renewbibmacro{in:}{}

\citetrackerfalse
\DefineBibliographyStrings{english}{%
  mathesis = {MSc thesis},
}

\usepackage[title]{appendix}

\usepackage{authblk}

\usepackage[hang]{footmisc}
\usepackage[ruled, algo2e]{algorithm2e} 
\usepackage{algpseudocode}
\SetKw{KwInitialization}{Initialize}
\SetKw{KwInput}{Input:}
\SetKw{KwOutput}{Output:}

\usepackage{booktabs}
\title{Exjobb}
\begin{document}

\title{ {\sffamily A similarity-based Bayesian mixture-of-experts model} }
\author[$*$\textsf{,}$\dagger$\textsf{,}$\ddagger$]{Tianfang Zhang}
\author[$\dagger$]{Rasmus Bokrantz}
\author[$*$]{Jimmy Olsson}
{
\affil[$*$]{Department of Mathematics, KTH Royal Institute of Technology, Stockholm SE-100 44, Sweden}
\affil[$\dagger$]{RaySearch Laboratories, Eugeniavägen 18, Solna, Stockholm SE-171 64, Sweden}
\affil[$\ddagger$]{Silo AI, Fredrikinkatu 57 C, Helsinki FI-00100, Finland}
}
\date{\textsf{August 3, 2022}}
\maketitle

\begin{quote}
{\centering
\section*{Abstract}
}
We present a new nonparametric mixture-of-experts model for multivariate regression problems, inspired by the probabilistic $k$-nearest neighbors algorithm. Using a conditionally specified model, predictions for out-of-sample inputs are based on similarities to each observed data point, yielding predictive distributions represented by Gaussian mixtures. Posterior inference is performed on the parameters of the mixture components as well as the distance metric using a mean-field variational Bayes algorithm accompanied with a stochastic gradient-based optimization procedure. The proposed method is especially advantageous in settings where inputs are of relatively high dimension in comparison to the data size, where input--output relationships are complex, and where predictive distributions may be skewed or multimodal. Computational studies on five datasets, of which two are synthetically generated, illustrate clear advantages of our mixture-of-experts method for high-dimensional inputs, outperforming competitor models both in terms of validation metrics and visual inspection. 
\newline
\begin{spacing}{0.9}
{\sffamily\small \noindent \textbf{Keywords:} Mixture-of-experts, nonparametric Bayesian regression, $k$-nearest neighbors, pseudolikelihood, variational inference, reparameterization trick.}
\end{spacing}
\end{quote}

\tolerance=1000

\section{Introduction}
\label{introduction}

Large parts of contemporary research in machine learning presume the availability of massive datasets and focus on pure prediction. In problems where data is complex as well as scarce, however, and where it is of high importance to quantify uncertainties of predictions, rigorous probabilistic modeling is essential for achieving good generalization performance. The Bayesian formalism provides a natural setting for such modeling---by specifying a model $p(\{y^n\}_n, y^*)$ jointly over observed data $\{y^n\}_n \subset \mathcal{Y}$ and future data $y^* \in \mathcal{Y}$, denoting by $\mathcal{Y}$ the space of outputs, obtaining the predictive distribution $p(y^* \mid \{y^n\}_n)$ reduces to an integration problem. When using parametric models, prior assumptions on the parameters act as regularizations preventing overfitting, but the requirement of an explicit parameterization may be problematic when outputs depend on inputs through highly nonlinear relationships. Nonparametric alternatives, on the other hand, are associated with other issues---simple models such as Gaussian processes struggle with multivariate, skewed and multimodal outputs, and more sophisticated nonparametric Bayesian regression models based on Dirichlet processes require \emph{ad hoc} constructions to incorporate input dependency and specially developed Markov chain Monte Carlo algorithms for posterior inference. To address these challenges, in this paper, we propose a novel nonparametric mixture-of-experts model designed to be able to model any predictive density while being robust against small datasets and overfitting, equipped with a mean-field variational Bayes algorithm for posterior inference. 

For modeling irregular densities, Gaussian mixtures are a popular choice as this class of distributions is dense in the set of all probability distributions \citep{nguyen}, thus being able to approximate any target density arbitrarily well. Parameter inference for finite Gaussian mixtures is particularly tractable with algorithms such as expectation--maximization (EM), mean-field variational Bayes and Gibbs sampling \citep{murphy}. In a supervised learning context, if a discriminative approach is preferred, an extension from mixture models are mixture-of-experts models, first introduced by \citet{jacobsjordan}, which allow the mixture components---the experts---as well as the mixture weights---the gate---to depend on inputs $x \in \mathcal{X}$ from an input space $\mathcal{X}$. A simple example is the model
\begin{align*}
z \mid x, \theta &\sim \operatorname{Cat}\!\left( \operatorname{softmax}(\Phi x) \right), \\
y \mid x, z = c, \theta &\sim \operatorname{N}(\mu_c, \Sigma_c),
\end{align*}
where $\theta = (\{(\mu_c, \Sigma_c)\}_c, \Phi)$, $\Phi$ is some transformation matrix and $z$ is a categorical latent variable with probability masses given by the softmax of $\Phi x$. Note that here, only the gate is input-dependent, which is sometimes known as a gating network mixture-of-experts model \citep{murphymurphy}---however, as a more complex example, we could also let the experts be input-dependent by e.g. letting the $\mu_c$ and $\Sigma_c$ be defined through parameterized transformations of $x$, known as a full mixture-of-experts model. More advanced designs have been investigated in the literature---for example, \citet{jordanjacobs} and \citet{bishopsvensen} use a hierarchical gating network for increased model complexity, and \citet{xu}, \citet{ueda}, \citet{baldacchino} and \citet{ingrassia} use a gate which itself is Gaussian or mixture-Gaussian. While some of these \citep{jordanjacobs, xu} use EM for maximum-likelihood parameter inference, others \citep{waterhouse, ueda, bishopsvensen, baldacchino} have employed variational Bayesian methods to approximate a posterior over the parameters, thereby reducing the imminent risk of overfitting and collapsing variances associated with maximum-likelihood \citep{bishop, waterhouse}. For an extensive review of various forms of mixture-of-experts models and methods for parameter inference, see \citet{yuksel}.

Common for the above models is that they are parametric, assuming conditional independence across the data pairs $\{(x^n, y^n)\}_n \subset \mathcal{X} \times \mathcal{Y}$ given some finite-dimensional parameter $\theta$. For exchangeable data, such a model is guaranteed to exist without loss of generality in the infinite-data limit by de Finetti's theorem \citep{schervish}. However, due to the form of the posterior predictive distribution of the output $y^* \in \mathcal{Y}$ corresponding to a new input $x^* \in \mathcal{X}$, written as
\[
p(y^* \mid x^*, \{(x^n, y^n)\}_n) = \operatorname{\mathbb{E}}^{p(\theta \; \mid \; \{(x^n, y^n)\}_n)} p(y^* \mid x^*, \theta)
,\]
all knowledge learned from the training data must be encapsulated in the posterior $p(\theta \mid \{(x^n, y^n)\}_n)$. While this does not present any practical limitations for most types of datasets, problems typically arise when the relationships between inputs and outputs are complex, when the dimension $\operatorname{dim} \mathcal{X}$ of the inputs is medium-high or high, and when data is scarce---in general, the former two situations force $\theta$ to also be of relatively high dimension which, in combination with the latter, renders inference especially hard. Nonparametric models, in contrast, characterized by their growth in complexity along with increasing data availability and often being based on comparing similarities between data points rather than learning a mapping from input to output, do not suffer from the same problem. Such similarity-based models are most easily understood from a predictive perspective---when inferring $y^*$ given $x^*$ and training data $\{(x^n, y^n)\}_n$, we first determine similarities between $x^*$ and each $x^n$ and base our prediction on these similarities and the $y^n$. The simplest example is the $k$-nearest neighbors algorithm, which has been extended to a probabilistic framework by \citet{holmes}, who replace the pure averaging over the $k$-nearest neighbors by a softmax weighting and sample from the posterior of $k$ and the softmax sharpness parameter using Markov chain Monte Carlo (MCMC) methods. The approach has since been revisited and refined by \citet{cucala} and, subsequently, \citet{friel}. For regression problems, we have similar models such as kernel machines \citep{bishop} and the more probabilistically rigorous Gaussian processes \citep{rasmussenwilliams}. Although several extensions of Gaussian processes \citep{bonilla, rasmussenghahramani} have been proposed to address, for example, their limitation to univariate outputs and Gaussian likelihoods, no combined method yet exists for handling arbitrary predictive distributions. 

Other nonparametric Bayesian methods for conditional density estimation involve Dirichlet process mixture models (DPMMs) \citep{ferguson, sethuraman, antoniak}. Here, the Dirichlet process is used as a nonparametric prior in an infinite mixture model
\begin{align*}
G &\sim \operatorname{DP}(\alpha, G_0), \\
\theta \mid G &\sim G, \\
y \mid \theta &\sim p(y \mid \theta),
\end{align*}
where $\operatorname{DP}(\alpha, G_0)$ is the Dirichlet process with concentration parameter $\alpha$ and base measure $G_0$ and $p(y \mid \theta)$ refers to some likelihood model. Several methods exist for using the DPMM in a regression setting, which has been reviewed by \citet{muller_book}. In particular, while simpler approaches are based on estimating the regression means and residuals separately, in a fully nonparametric regression method, the dependency on an input $x$ may be introduced by replacing $G$ with an $x$-indexed analogue $G_x$ and instead placing the nonparametric prior over $\{G_x\}_{x \in \mathcal{X}}$. Known as the dependent Dirichlet process (DDP) \citep{maceachern}, this model has been investigated with various choices of priors for $\{G_x\}_{x \in \mathcal{X}}$---for example, \citet{deiorio_anova} proposed the ANOVA-DDP model in which locations in the stick-breaking representation of the Dirichlet process \citep{sethuraman} are linear in the inputs, which was further developed into the linear DDP model \citep{deiorio_linear}; \citet{dunson} proposed a kernelized stick-breaking process in which the breaking probabilities are input-dependent and similarity-based; \citet{jara} instead proposed to replace the usual beta priors in the stick-breaking representation of the Dirichlet process by logistic-transformed Gaussian processes, deviating from the DPMM. Common drawbacks of these models, however, are that they require the modeling of an input-to-parameter process $G_x$, which is in general multivariate, that the construction of a prior over $\{G_x\}_{x \in \mathcal{X}}$ essentially amounts to yet another multivariate regression problem, and that the available posterior MCMC inference algorithms exploit the respective specific forms of the priors. For instance, if using a normal likelihood $p(y \mid \theta) = \operatorname{N}(y \mid \mu, \Sigma)$ with $\theta = (\mu, \Sigma)$, it is not clear how one would go about specifying the $(\operatorname{dim} \mathcal{Y} + (\operatorname{dim} \mathcal{Y})^2)$-dimensional process $G_x$ and its prior in such a way that the input-to-parameter relation is adequately modeled and posterior inference is feasible. As an alternative perspective, known as the conditional DPMM \citep{muller_cdpmm, cruz-marzelo}, one may model the data generatively and fit a DPMM to the set $\{(x^n, y^n)\}_n \subset \mathcal{X} \oplus \mathcal{Y}$ of concatenated input--output pairs, from which conditional distributions are easily obtained due to the mixture-Gaussian form of the likelihood. While MCMC methods, which are known to scale poorly with model size and data dimensionality, are used for posterior inference in all previous cases, an alternative variational Bayes algorithm \citep{bleijordan} is possible in the this model. On the other hand, the generative approach of the conditional DPMM relies on modeling both the input and output spaces using a mixture model, which may be unsuitable if the dimensionality of the inputs is large compared to that of the outputs. Moreover, the conditional DPMM relies on an approximate independence assumption across the data \citep{muller_book}, and the model is therefore not similarity-based in the sense that $x^*$ is compared with each $x^n$ when predicting $y^*$.

In this paper, we present a new similarity-based mixture-of-experts model combining the advantages of similarity-based nonparametric models with the flexibility of Gaussian mixtures, all while avoiding generative modeling and MCMC posterior inference. Specifically, the model uses multivariate Gaussian experts and a gating network comprising two layers of transitions when making predictions: a first layer where similarities between $x^*$ and each $x^n$ are used to compute transition probabilities, and a second layer where the conditional probabilities of belonging to each expert $c$ given each $y^n$ are computed. As such, in contrast to aforementioned DPMM-based nonparametric Bayesian regression methods, in which input-to-parameter relations must be explicitly modeled by $G_x$ and its prior, all input--output mappings are handled through the observed data pairs $\{(x^n, y^n)\}_n$. The model entails a predictive likelihood on the form of a multivariate Gaussian mixture, capable of modeling marginals as well as dependencies across the components of the output variable. Parameters are inferred using a mean-field variational Bayes algorithm, where local approximations of the likelihood are introduced to make variational posteriors analytically tractable and where the corresponding variational parameters may be updated using EM iterations. A gradient-based way of optimizing the similarity metric of the first transition is presented, which uses the reparameterization trick \citep{kingma} to make stochastic gradients available through Monte Carlo sampling. The result is an algorithm generally suitable for datasets for which input--output relationships are intricate, the number of observations is relatively small, the output variables are multivariate, the predictive distributions may be multimodal as well as skewed, and for which it is desirable to estimate the predictive distributions as exactly as possible. The proposed method was tested on two artificially generated datasets, a dataset from medical physics containing dose statistics from radiation therapy treatments, the California housing dataset \citep{pace} comprising geographic location and features of housing districts in California, and a dataset \citep{africa_soil} of soil functional properties and infrared spectroscopy measurements in Africa. In terms of both visual inspection and validation metrics, our model outperformed a conditional DPMM on all but the radiation therapy dataset, where it gave similar results, and performed consistently better than a Gaussian process baseline model. In particular, the experiments serve to illustrate the advantages of the proposed mixture-of-experts model on data with relatively high-dimensional inputs and irregularly shaped output distributions. 

The sequel of this paper is organized as follows: the assumptions of the model are accounted for in Section \ref{secmodel}, details of the mean-field variational Bayes algorithm are given in Section \ref{secparameterinference}, results from the computational study are presented in Section \ref{seccomputationalstudy} and discussed in Section \ref{secdiscussion}, and derivations of some key facts and other algorithm details are given in Appendices \ref{secappendix}, \ref{emememem}, \ref{appendixelbo}, \ref{appedix_stochastic_gate_opt} and \ref{hyperparameter_selection}.

\section{Model}
\label{secmodel}

Let $\{(x^n, y^n)\}_{n = 1}^N \subset \mathcal{X} \times \mathcal{Y}$ be exchangeable pairs of random variables consisting of inputs $x^n$ and outputs $y^n$, where $\mathcal{X}$ and $\mathcal{Y}$ are some appropriate vector spaces of random variables. Given observations of these data pairs, which are referred to as the training data, our main task is to be able to, for each new input $x^* \in \mathcal{X}$, obtain the predictive distribution $p(y^* \mid x^*, \{(x^n, y^n)\}_n)$ of the corresponding output $y^* \in \mathcal{Y}$. All random variables are modeled on a common probability space with probability measure $\operatorname{\mathbb{P}}$, using $p$ for densities or masses induced by $\mathbb{P}$. We will frequently use $p(\{z_i\}_{i=1}^I)$ as shorthand for the joint density $p(z_1, z_2, \dots, z_I)$ and $\operatorname{\mathbb{E}}^{p(z)}$ for the marginalization over a random variable $z \sim p(z)$. 

Before introducing our model, it will be instructive to first formulate a conventional kernel regression model using latent variables:
\begin{examp}[Nadaraya--Watson estimator]
\label{nadarayawatson}
Consider a model where prediction of $y^*$ given $x^*$ may be written as a noisy linear combination
\begin{equation}
\label{kernelregression}
y^* = \sum_n \frac{k(x^*, x^n)}{\sum_{n'} k(x^*, x^{n'})} y^n + \varepsilon^*
\end{equation}
of the training outputs $y^n$, where $\varepsilon^* \sim \operatorname{N}(0, \epsilon^2)$ is the regression error and $k : \mathcal{X}^2 \to \mathbb{R}$ is some positive definite kernel---this is sometimes known as the Nadaraya--Watson estimator \citep{bishop}. For simplicity, we will use the radial basis function kernel $k(x, x') = {\operatorname{exp}(-\Vert x - x' \Vert^2 / (2\ell^2))}$ with standard Euclidean norm $\Vert \cdot \Vert$ and lengthscale parameter $\ell$. We can now reformulate this using latent variables. In particular, let $u^*$ be a random variable supported on $\{1, \dots, N\}$ such that 
\[
p(u^* = n \mid x^*, \{(x^{n'}, y^{n'})\}_{n'}) \propto \operatorname{N}(x^* \mid x^n, \ell^2 I)
,\]
$I$ being the identity matrix, which may be interpreted as $u^*$ \say{choosing} one of the observed data points $n$ and setting up a normal distribution centered around $x^n$. If, conditional upon the choice $u^* = n$, the prediction of $y^*$ is $y^n$ with some normal uncertainty of variance $\epsilon^2$---that is, $p(y^* \mid u^* = n, x^*, \{(x^{n'}, y^{n'})\}_{n'}) = \operatorname{N}(y^* \mid y^n, \epsilon^2)$---we obtain
\[
p(y^* \mid x^*, \{(x^n, y^n)\}_n) = \sum_{n} \frac{\operatorname{N}(x^* \mid x^n, \ell^2 I)}{\sum_{n'} \operatorname{N}(x^* \mid x^{n'}, \ell^2 I)} \operatorname{N}(y^* \mid y^n, \epsilon^2)
.\]
From the radial basis form of $k$, it is easy to see that this is equivalent to the conventional kernel regression formulation (\ref{kernelregression}).
\end{examp}

\subsection{Main setup}

To assemble our model, we will start from a predictive perspective. We use $C$ normal distributions $\{\operatorname{N}(\mu_c, \Sigma_c)\}_{c=1}^C$ as experts, where the means $\mu_c$ and covariances $\Sigma_c$ are parameters not depending on the inputs. The gate then consists of two layers of transitions---one between the new input $x^*$ and each training input $x^n$, and one between each training output $y^n$ and each expert $c$---which are represented by, respectively, latent variables $u^*$ and $z^*$ supported on $\{1, \dots, N\}$ and $\{1, \dots, C\}$. In particular, we let 
\[
p(u^* = n \mid x^*, \{(x^{n'}, y^{n'})\}_{n'}, \theta) \propto \operatorname{N}(x^* \mid x^n, \Lambda^{-1})
,\]
where $\Lambda$ is a precision matrix (as a generalization from $\ell^{-2} I$ in Example \ref{nadarayawatson}), and
\[
p(z^* = c \mid u^* = n, x^*, \{(x^{n'}, y^{n'})\}_{n'}, \theta) \propto \operatorname{N}(y^n \mid \mu_c, \Sigma_c)
,\]
using $\theta = (\Lambda, \{(\mu_c, \Sigma_c)\}_c)$ for the collection of all parameters. Furthermore, we let
\[
p(y^* \mid z^* = c, u^* = n, x^*, \{(x^{n'}, y^{n'})\}_{n'}, \theta) = \operatorname{N}(y^* \mid \mu_c, \Sigma_c)
\]
for all $n$ and $c$. The complete-data predictive likelihood is then written as
\begin{equation}
\label{completedatapred}
\begin{split}
p(y^*, u^*, & \, z^* \mid x^*, \{(x^n, y^n)\}_n, \theta) \\
&= p(y^* \mid z^*, \theta) p(z^* \mid u^*, \{y^{n}\}_{n}, \theta) p(u^* \mid x^*, \{x^n\}_n, \theta)  \\
&= \prod_n \prod_c \Bigg( \frac{\operatorname{N}(x^* \mid x^n, \Lambda^{-1})}{\sum_{n'} \operatorname{N}(x^* \mid x^{n'}, \Lambda^{-1})} \\
&\quad\quad\quad\quad \times \frac{\operatorname{N}(y^n \mid \mu_c, \Sigma_c)}{\sum_{c'} \operatorname{N}(y^n \mid \mu_{c'}, \Sigma_{c'})} \operatorname{N}(y^* \mid \mu_c, \Sigma_c) \Bigg)^{1_{z^* \; = \; c} 1_{u^* \; = \; n}},
\end{split}
\end{equation}
leading to the observed-data predictive likelihood 
\begin{equation*}
\begin{split}
p(& y^* \mid x^*, \{(x^n, y^n)\}_n, \theta) \\
&= \sum_c \left( \sum_n \frac{\operatorname{N}(x^* \mid x^n, \Lambda^{-1})}{\sum_{n'} \operatorname{N}(x^* \mid x^{n'}, \Lambda^{-1})} \frac{\operatorname{N}(y^n \mid \mu_c, \Sigma_c)}{\sum_{c'} \operatorname{N}(y^n \mid \mu_{c'}, \Sigma_{c'})} \right) \operatorname{N}(y^* \mid \mu_c, \Sigma_c),
\end{split}
\end{equation*}
which is a mixture of Gaussians. The first and second fraction within the parentheses of the above display are recognized as probabilities associated with first and second transitions, respectively. The precision matrix $\Lambda$ naturally induces a Mahalanobis-form distance metric $d_{\mathcal{X}}(x, x') = (x - x')^{\operatorname{T}} \Lambda (x - x')$ on $\mathcal{X}$, whereby the first transitions are obtained as a softmax transformation of the vector $(-d_{\mathcal{X}}(x^*, x^n) / 2)_n$. Note, in particular, that this is a gating network mixture-of-experts model as the mean--covariance pairs $(\mu_c, \Sigma_c)$ are not input-dependent. This is a deliberate choice in light of our preference to avoid stipulating an input-to-parameter map, which is in general a high-dimensional and potentially complex transformation, as discussed in Section \ref{introduction}---instead, we let all input--output transitions occur through the observed data pairs $\{(x^n, y^n)\}_n$.

Having established the predictive pipeline, we now face a similar problem as \citet{holmes} originally did in the probabilistic $k$-nearest neighbors context of constructing a joint complete-data likelihood $p(\{(y^n, u^n, z^n)\}_n \mid \{x^n\}_n, \theta)$ given the training inputs. It is intuitively reasonable to require that the full conditionals $p(y^n, u^n, z^n \mid x^n, \{(x^{n'}, y^{n'})\}_{n' \neq n}, \theta)$ have forms analogous to the predictive likelihood (\ref{completedatapred}). As pointed out by \citet{cucala}, however, directly defining the joint likelihood as the product of all full conditionals as in \citet{holmes} will lead to an improperly normalized density. Noting that the asymmetry of the neighborhood relationships is the main cause of difficulties in specifying a well-defined model, \citet{cucala} instead use a symmetrized version and view the training data as Markov random field \citep{murphy}, defining a joint distribution up to a normalizing constant corresponding to a Boltzmann-type model. The idea is reused in \citet{friel}, who replace the arguably superficial symmetrized neighborhood relationships with distance metrics. Both papers present a pseudolikelihood approximation \citep{besag} of the joint distribution as a possible approach, although focus is reserved for other methods of handling the intractable normalizing constant. 

Historically, there has been considerable debate regarding the use of a proper joint likelihood versus a conditionally specified model, for which a joint likelihood may not even exist \citep{besagkooperburg, cucala}. As argued by \citet{besag}, the appeal of the latter choice is mainly due to the increased interpretability of specifying relationships directly through full conditionals, whereby the pseudolikelihood is always available as a viable surrogate for the joint likelihood. Thus, despite a certain lack of statistical coherence, this methodology allows the translation of any predictive pipeline, such as that of the $k$-nearest neighbor algorithm, into a probabilistic framework. Subsequent investigations of the model by \citet{holmes} gave different motivations for using this approach---\citet{manocha} noticed the resemblance of parameter inference in such a model with leave-one-out cross-validation, whereas \citet{yoon} explained that the Boltzmann-type model by \citet{cucala} would lead to difficulties due to the intractable normalizing constant. The interpretation of using such a pseudolikelihood is that dependencies between the data points beyond first order are ignored \citep{friel}, which may be a reasonable simplification in many applications.

Given this, we settle on the pseudolikelihood approach and proceed to define the full conditionals of our model as
\begin{align*}
p(y^n, u^n, & \, z^n \mid x^n, \{(x^{n'}, y^{n'})\}_{n' \neq n}, \theta) \\
&= \prod_{n' \neq n} \prod_c \Bigg( \frac{\operatorname{N}(x^n \mid x^{n'}, \Lambda^{-1})}{\sum_{n''} \operatorname{N}(x^n \mid x^{n''}, \Lambda^{-1})} \\
&\quad\quad\quad\quad \times \frac{\operatorname{N}(y^{n'} \mid \mu_c, \Sigma_c)}{\sum_{c'} \operatorname{N}(y^{n'} \mid \mu_{c'}, \Sigma_{c'})} \operatorname{N}(y^n \mid \mu_c, \Sigma_c) \Bigg)^{1_{z^n \; = \; c} 1_{u^n \; = \; n'}},
\end{align*}
recognized as the analogue of (\ref{completedatapred}), where the latent variables $u^n$, $z^n$ are now supported on $\{1, \dots, N\} \setminus \{n\}$ and $\{1, \dots, C\}$, respectively. The joint complete-data pseudolikelihood is then written as
\begin{equation}
\label{pseudolikelihood}
p_{\mathrm{pseudo}}(\{(y^n, u^n, z^n)\}_n \mid \{x^n\}_n, \theta) = \prod_n p(y^n, u^n, z^n \mid x^n, \{(x^{n'}, y^{n'})\}_{n' \neq n}, \theta)
\end{equation}
---while it is important to remember that this is not a proper joint likelihood, we will drop the subscript in the following for convenience of notation. To complete our Bayesian model, we assume that the precision matrix $\Lambda$ and the mean--covariance pairs $(\mu_c, \Sigma_c)$ are \emph{a priori} independent with Wishart and normal--inverse-Wishart distributions, respectively. This is written as
\[
p(\theta) = \operatorname{W}(\Lambda \mid \Lambda_0, \eta_0) \prod_c \operatorname{N}\!\left(\mu_c \; \bigg| \; \mu_0, \frac{1}{\kappa_0} \Sigma_c\right) \operatorname{IW}(\Sigma_c \mid \Sigma_0, \nu_0) 
,\]
where $\Lambda_0$, $\eta_0$, $\mu_0$, $\kappa_0$, $\Sigma_0$ and $\nu_0$ are hyperparameters. In order to ensure that the distributions are well-defined, we must require that $\Lambda_0$ and $\Sigma_0$ be positive definite, that $\eta_0 > \operatorname{dim} \mathcal{X} - 1$ and $\nu_0 > \operatorname{dim} \mathcal{Y} - 1$, and that $\kappa_0 > 0$ \citep{anderson}. The discussion of general-purpose default methods to set hyperparameters is deferred to Appendix \ref{hyperparameter_selection}---in particular, the number $C$ of experts should generally be high to allow flexibility in the set of possible predictive distributions. Using (\ref{pseudolikelihood}), this leads to the log-pseudojoint 
\begin{equation}
\begin{split}
\label{logjoint}
\log &\, p(\{(y^n, u^n, z^n)\}_n, \theta \mid \{x^n\}_n) \\
&= \log \operatorname{W}(\Lambda \mid \Lambda_0, \eta_0) + \sum_c \left( \log \operatorname{N}\!\left(\mu_c \; \bigg| \; \mu_0, \frac{1}{\kappa_0} \Sigma_c\right) + \log \operatorname{IW}(\Sigma_c \mid \Sigma_0, \nu_0) \right) \\
&\quad + \sum_n \sum_{n' \neq n} \sum_c 1_{z^n \; = \; c} 1_{u^n \; = \; n'} \Bigg(  \log \operatorname{N}(y^n \mid \mu_c, \Sigma_c) \\
&\quad\quad\quad + \log \operatorname{N}(y^{n'} \mid \mu_c, \Sigma_c) - \log\!\Bigg( \sum_{c'} \operatorname{N}(y^{n'} \mid \mu_{c'}, \Sigma_{c'}) \Bigg) \\
&\quad\quad\quad + \log \operatorname{N}(x^n \mid x^{n'}, \Lambda^{-1}) - \log\!\Bigg( \sum_{n'' \neq n} \operatorname{N}(x^n \mid x^{n''}, \Lambda^{-1}) \Bigg) \Bigg).
\end{split}
\end{equation}

\section{Parameter inference}
\label{secparameterinference}

\subsection{Mean-field variational Bayes}

Motivated by the Gaussian mixture form of the full conditionals, we will base our parameter inference on a mean-field variational Bayes algorithm. In general, given some observed variables $y$ and hidden variables $z$, the main idea of variational inference is to approximate an intractable posterior $p(z \mid y)$ by a variational posterior $q(z)$ from some family $\mathcal{Q}$ of distributions. The optimal $q^*(z) \in \mathcal{Q}$ is found by minimizing the Kullback--Leibler (KL) divergence $d_{\operatorname{KL}}(q \; \Vert \; p(\cdot \mid y))$ according to 
\begin{equation*}
\begin{split}
q^*(z) &= \; \underrel{\operatorname{arg\,min}}{q(z) \in \mathcal{Q}} \; -\operatorname{\mathbb{E}}^{q(z)} \log\!\left( \frac{p(z \mid y)}{q(z)} \right) = \; \underrel{\operatorname{arg\,max}}{q(z) \in \mathcal{Q}} \; \operatorname{ELBO}(q),
\end{split}
\end{equation*}
where the evidence lower bound (ELBO) is defined as 
\[
\operatorname{ELBO}(q) = \operatorname{\mathbb{E}}^{q(z)} \log\!\left( \frac{p(y, z)}{q(z)} \right)
,\]
its name originating from the fact that the bound $\log p(y) \geq \operatorname{ELBO}(q)$ always holds. By mean-field variational inference, we mean using as $\mathcal{Q}$ the family of all distributions on the fully factorized form
\[
q(z) = \prod_i q(z_i)
,\]
where we have block-decomposed $z$ as $z = (z_i)_i$. In this case, one can show that the optimal distribution $q^*(z_i)$ for each component $i$ has log-density equal up to an additive constant to the log-joint with $\{z_{i'}\}_{i' \neq i}$ marginalized out---that is,
\[
\log q^*(z_i) \stackrel{\mathrm{c}}{=} \operatorname{\mathbb{E}}^{q(\{z_{i'}\}_{i' \neq i})} \log p(y, z) 
.\]
This is used to construct a coordinate ascent algorithm, where a sequence $\{q^l\}_{l \geq 0}$ of distributions are devised to approximate $q$ increasingly better and updates are performed according to the above by, at each $l$, iterating through the components of $z$ and taking expectations always with respect to the latest versions of the $q(z_i)$. For a detailed review of variational inference methods, see \citet{blei}.

In our case, counting both the latent variables $u^n$, $z^n$ and the parameters $\theta$ as hidden variables, we propose the factorization
\[
q(\theta, \{(u^n, z^n)\}_n) = q(\Lambda) q(\{(\mu_c, \Sigma_c)\}_c) q(\{(u^n, z^n)\}_n)
.\]
An issue with the terms containing logarithms of summed normal densities in (\ref{logjoint}) is that they do not lead to tractable variational posteriors. \citet{xu} proposed a generative model, which was also used by \citet{baldacchino}, where each $p(x^n)$ was on the form of a Gaussian mixture, leading to analytically solvable variational posteriors at the cost of introducing additional parameters to the model and thus also accompanying hyperparameters. Trying to avoid this, we settle for a slightly different approach and instead approximate the log-sum-exp function $\operatorname{LSE}$ by the linearization 
\begin{equation}
\label{lseapprox}
\operatorname{LSE}(\xi) = \log\!\left( \sum_i e^{\xi_i} \right) \approx \log\!\left( \sum_i e^{\eta_i} \right) + \sum_i \frac{e^{\eta_i}}{\sum_{i'} e^{\eta_{i'}}} (\xi_i - \eta_i),
\end{equation}
where $\xi = (\xi_i)_i$ and $\eta = (\eta_i)_i$. In particular, we let $s = (s_{nc})_{n, c}$ and $t = (t_{nn'})_{n, n' \neq n}$ be the softmax-transformed locations of the linearizations---that is, for each $n$, we have
\begin{equation}
\label{secondtransapprox}
\log\!\left( \sum_c \operatorname{N}(y^n \mid \mu_c, \Sigma_c) \right) \stackrel{\mathrm{c}}{\approx} \sum_c s_{nc} \log \operatorname{N}(y^n \mid \mu_c, \Sigma_c)
\end{equation}
and
\begin{equation}
\label{firsttransapprox}
\log\!\left( \sum_{n' \neq n} \operatorname{N}(x^n \mid x^{n'}, \Lambda^{-1}) \right) \stackrel{\mathrm{c}}{\approx} \sum_{n' \neq n} t_{nn'} \log \operatorname{N}(x^n \mid x^{n'}, \Lambda^{-1}).
\end{equation}
Here, $s_n = (s_{nc})_c$ and $t_n = (t_{nn'})_{n' \neq n}$, and $\stackrel{\mathrm{c}}{\approx}$ denotes approximate equality up to a constant not depending on $\theta$. We regard $s$ and $t$ as variational parameters---the presentation of how they may be optimized is deferred to Section \ref{optvarparams}. It will be shown that, provided that the values of $s$ and $t$ are appropriate, our model will become conditionally conjugate under the assumptions above, resulting in variational posteriors from the same parametric families as our priors.

Letting $\{q^l\}_{l \geq 0}$ be a sequence of variational posterior approximations, we use the assumed factorization to obtain update equations according to
\[
\log q^{l+1}(\{(u^n, z^n)\}_n) \stackrel{\mathrm{c}}{=} \operatorname{\mathbb{E}}^{q^l(\Lambda) q^l(\{(\mu_c, \Sigma_c)\}_c)} \log p(\{(y^n, u^n, z^n)\}_n, \theta \mid \{x^n\}_n)
,\]
\[
\log q^{l+1}(\{(\mu_c, \Sigma_c)\}_c) \stackrel{\mathrm{c}}{=} \operatorname{\mathbb{E}}^{q^{l+1}(\{(u^n, z^n)\}_n) q^l(\Lambda)} \log p(\{(y^n, u^n, z^n)\}_n, \theta \mid \{x^n\}_n)
\]
and
\[
\log q^{l+1}(\Lambda) \stackrel{\mathrm{c}}{=} \operatorname{\mathbb{E}}^{q^{l+1}(\{(u^n, z^n)\}_n) q^{l+1}\{(\mu_c, \Sigma_c)\}_c)} \log p(\{(y^n, u^n, z^n)\}_n, \theta \mid \{x^n\}_n)
.\]
These will be referred to as the E-step, the M-step I and the M-step II, respectively, due to the strong resemblance with the standard EM algorithm for fitting Gaussian mixtures \citep{murphy} and the expectation--conditional maximization algorithm \citep{rubin}. Let also $s^l$, $t^l$ be the values of the variational parameters $s$ and $t$ at iteration $l$, and let
\[
\omega_{c, nn'}^l = q^l(z^n = c, u^n = n'), \quad \Omega_{nn'}^l = \sum_c \omega_{c, nn'}^l
,\]
for each $c$, $n$ and $n'$. We can then summarize the details of the updates in Propositions \ref{estep}, \ref{mstep1} and \ref{mstep2} below, for which derivations are given in Appendix \ref{secappendix}. The overall algorithm, including the steps described in Section \ref{optvarparams}, is outlined in Algorithm \ref{algo}.

\begin{prop}[Variational posterior, E-step]
\label{estep}
For all $l$, we have the decomposition
\[
q^l(\{(u^n, z^n)\}_n) = \prod_n q^l(u^n, z^n) = \prod_n \prod_{n' \neq n} \prod_c \left(\omega_{c, nn'}^l \right)^{1_{z^n \; = \; c} 1_{u^n \; = \; n'}}
.\]
Updates are given by
\begin{align*}
\log &\, \omega_{c, nn'}^{l+1} \\
&\stackrel{\mathrm{c}}{=} \sum_{i=1}^{\dim \mathcal{Y}} \left( \psi\!\left( \frac{\nu_c^l + 1 - i}{2} \right) - \frac{1}{2}\sum_{c'} s_{n'c'}^l \psi\!\left( \frac{\nu_{c'}^l + 1 - i}{2} \right) \right) \\
&\quad\quad -\log \det \Sigma_c^l + \frac{1}{2} \sum_{c'} s_{n'c'}^l \log \det \Sigma_{c'}^l + \dim \mathcal{Y} \left( -\frac{1}{\kappa_c^l} + \frac{1}{2} \sum_{c'} s_{n'c'}^l \frac{1}{\kappa_{c'}^l} \right) \\
&\quad\quad -\frac{1}{2} \nu_c^l (y^n - \mu_c^l)^{\operatorname{T}} \Sigma_c^{-l} (y^n - \mu_c^l) -\frac{1}{2} \nu_c^l (y^{n'} - \mu_c^l)^{\operatorname{T}} \Sigma_c^{-l} (y^{n'} - \mu_c^l) \\
&\quad\quad + \frac{1}{2} \sum_{c'} s_{n'c'}^l (y^{n'} - \mu_{c'}^l)^{\operatorname{T}} \Sigma_{c'}^{-l} (y^{n'} - \mu_{c'}^l) \\ 
&\quad\quad -\frac{1}{2} \eta^l (x^n - x^{n'})^{\operatorname{T}} \Lambda^l (x^n - x^{n'}) + \frac{1}{2} \eta^l \sum_{n'' \neq n} t_{nn''}^l (x^n - x^{n''})^{\operatorname{T}} \Lambda^l (x^n - x^{n''}),
\end{align*}
where $\psi$ is the digamma function. 
\end{prop}

\begin{rem}
The values of $\omega_{c, nn'}^l$ may be found using the fact that $\sum_c \sum_{n' \neq n} \omega_{c, nn'}^l = 1$.
\end{rem}

\begin{prop}[Variational posterior, M-step I]
\label{mstep1}
For all $l$, we have the decomposition
\[
q^l(\{(\mu_c, \Sigma_c)\}_c) = \prod_c q^l(\mu_c \mid \Sigma_c) q^l(\Sigma_c)
,\]
with $q^l(\mu_c \mid \Sigma_c) = \operatorname{N}\!\left(\mu_c \mid \mu_c^l, \kappa_c^{-l} \Sigma_c \right)$ and $q^l(\Sigma_c) = \operatorname{IW}(\Sigma_c \mid \Sigma_c^l, \nu_c^l)$. Defining
\[
r_{nc}^l = \sum_{n' \neq n} \left( \omega_{c, nn'}^l + \omega_{c, n'n}^l - s_{nc}^l \Omega_{n'n}^l \right), \quad  R_c^l = \sum_n r_{nc}^l
,\]
updates are given by
\[
\kappa_c^{l+1} = \kappa_0 + R_c^{l+1}, \quad \mu_c^{l+1} = \frac{\kappa_0 \mu_0 + \sum_{n} r_{nc}^{l+1} y^n }{\kappa_c^{l+1}}
\]
and
\[
\nu_c^{l+1} = \nu_0 + R_c^{l+1}, \quad \Sigma_c^{l+1} = \Sigma_0 + \kappa_0 \mu_0 \mu_0^{\operatorname{T}} - \kappa_c^{l+1} \mu_c^{l+1} \mu_c^{(l+1)\mathrm{T}} + \sum_n r_{nc}^{l+1} y^n y^{n\mathrm{T}}.
\]
Moreover, each $\Sigma_c^l$ is positive definite if $r_{nc}^l > 0$ for all $n$.
\end{prop}

\begin{prop}[Variational posterior, M-step II]
\label{mstep2}
For all $l$, we have
\[
q^l(\Lambda) = \operatorname{W}(\Lambda \mid \Lambda^l, \eta_0)
,\]
and updates are given by
\[
\Lambda^{-(l+1)} = \Lambda_0^{-1} + \sum_n \sum_{n' \neq n} (\Omega_{nn'}^{l+1} - t_{nn'}^{l+1})(x^n - x^{n'})(x^n - x^{n'})^{\operatorname{T}}
.\]
\end{prop}

\subsection{Optimization of variational parameters}
\label{optvarparams}

\subsubsection{Linear programming for M-step I}
\label{optlinprog}

Proposition \ref{mstep1} provides a sufficient condition for all scale matrices $\Sigma_c^l$ to be positive definite, which is required for the inverse-Wishart distribution to be well-defined. We shall combine this with the local variational approximation method outlined in \citet{bishop} and \citet{watanabe} to set up an optimization problem to be solved at each iteration for updating each row $s_n$ in the second transition variational parameters $s$.

Recall, in particular, that the purpose of introducing $s$ and the associated linearization (\ref{secondtransapprox}) is to render variational posteriors analytically tractable through a mean-field approximation. We would like to optimize the accuracy of the approximation by maximizing the linearized complete-data pseudolikelihood $p(\{(y^n, u^n, z^n)\}_n \mid \{x^n\}_n, \theta)$, which depends on $s$. Although the exact solution is trivial with $\theta$ fixed, since $\theta$ is uncertain, we instead resort to marginalizing out $\theta$ according to the latest variational posterior $q^l(\theta)$ and maximize the corresponding log-pseudoevidence
\[
\log \operatorname{\mathbb{E}}^{q^l(\theta)} p(\{(y^n, u^n, z^n)\}_n \mid \{x^n\}_n, \theta)
.\]
The form of this objective motivates the use of an EM algorithm. These EM updates will be interleaved with the variational Bayes updates, similarly to the scheme in \citet{bishopsvensen}, using the previous iterate as initialization and performing a single iteration at each $l$. In addition, we will constrain each update so that the M-step is taken constrained such that the condition $r_{nc}^{l+1} > 0$ from Proposition \ref{mstep1} holds. Removing terms not depending on $s$ and noting that the problem separates in such a way that each $s_n$ can be optimized separately, we can devise an update equation where $s^{l+1}$ is set to the solution of the linear program
\begin{equation}
\label{secondtransopt}
\begin{aligned}
&\underset{s_n \; : \; s_n \geq 0, \; 1^{\operatorname{T}} s_n = 1}{\text{minimize}} 
&& \sum_c s_{nc} \operatorname{\mathbb{E}}^{q^l(\mu_c, \Sigma_c)} \log \operatorname{N}(y^n \mid \mu_c, \Sigma_c), & \\
& \quad\; \text{subject to} && s_{nc} \leq \frac{\sum_{n' \neq n}(\omega_{c, nn'}^l + \omega_{c, n'n}^l)}{\sum_{n' \neq n} \Omega_{n'n}^l} & \text{for each } c,
\end{aligned}
\end{equation}
with 
\begin{equation*}
\begin{split}
\operatorname{\mathbb{E}}^{q^l(\mu_c, \Sigma_c)} \log \operatorname{N}(y^n \mid \mu_c, \Sigma_c) &= \frac{1}{2} \sum_{i=1}^{\dim \mathcal{Y}} \psi\!\left( \frac{\nu_c^l + 1 - i}{2} \right) - \frac{1}{2} \log \det \Sigma_c^l  \\
&\quad\quad - \frac{1}{2} \frac{\dim \mathcal{Y}}{\kappa_c^l} - \frac{1}{2} \nu_c^l (y^n - \mu_c^l)^{\operatorname{T}} \Sigma_c^{-l} (y^n - \mu_c^l).
\end{split}
\end{equation*}
A detailed motivation for (\ref{secondtransopt}) is given in Appendix \ref{emememem}. Note that since
\[
\sum_c \frac{\sum_{n' \neq n}(\omega_{c, nn'}^l + \omega_{c, n'n}^l)}{\sum_{n' \neq n} \Omega_{n'n}^l} = \frac{1 + \sum_{n' \neq n} \Omega_{n'n}^l}{\sum_{n' \neq n} \Omega_{n'n}^l} > 1
,\]
the constraints are always feasible.

\subsubsection{Stochastic gradient method for M-step II}
\label{stochgrad}

For the first transition variational parameters $t$, note that unlike Proposition \ref{mstep1}, Proposition \ref{mstep2} does not come with a guarantee that the double sum added to $\Lambda_0^{-1}$ is positive definite---thus, the Wishart distribution constituting the variational posterior of $\Lambda$ may not be well-defined. A solution is to use the unconstrained analogue
\begin{equation}
\label{firsttransopt}
\begin{aligned}
&\underset{t_n \; : \; t_n \geq 0, \; 1^{\operatorname{T}} t_n = 1}{\text{minimize}} 
&& \sum_{n' \neq n} t_{nn'} \operatorname{\mathbb{E}}^{q^l(\Lambda)} \log \operatorname{N}(x^n \mid x^{n'}, \Lambda^{-1}) \\
\end{aligned}
\end{equation}
of the optimization problem (\ref{secondtransopt}), where, if the resulting inverse scale matrix $\Lambda^{-(l+1)}$ is not positive definite, its eigenvalues will be thresholded to be at least some small positive number. This is, however, somewhat unsatisfactory in that the eigenvalue thresholding step would become another layer of indirection besides from introducing $t$. An alternative method would be to instead specify beforehand, motivated by Proposition \ref{mstep2}, that $q^l(\Lambda) = \operatorname{W}(\Lambda \mid \Lambda^l, \eta_0)$ and find the variational parameter $\Lambda^l$ by maximizing $\operatorname{ELBO}(q^l)$ without the linearization of the log-sum-exp expression associated with the first transition, according to the generic approach described in \citet{blei}. While it is always possible to approximate analytically intractable expectations by Monte Carlo methods, since the ELBO is an expectation taken with respect to $q^l$ itself, one will in general need additional manipulations to obtain the corresponding estimate of the gradient. Various methods for such situations exist, examples including the score function gradient, which works for very general cases and can be used in conjunction with variance reduction techniques such as Rao--Blackwellization and control variates \citep{ranganath}, and the reparameterization trick \citep{kingma}.

Since $q^l(\Lambda)$ depends on $\Lambda^l$ continuously, we will use the reparameterization trick, which is understood to be the preferred method in such cases due to its lower variance compared to that of the score function gradient, even when using variance reduction \citep{kucukelbir}. With the Cholesky factorization $\Lambda^l = L^l L^{l\mathrm{T}}$, we will reparameterize according to the Bartlett decomposition \citep{anderson}, which states that $\Lambda \sim \operatorname{W}(\Lambda^l, \eta_0)$ may be written as
\[
\Lambda = L^l A A^{\operatorname{T}} L^{l \mathrm{T}}
,\]
where $A = (A_{ii'})_{i, i' = 1}^{\operatorname{dim} \mathcal{X}}$ is a lower triangular random matrix with all independent entries such that $A_{ii}^2 \sim \chi^2_{\eta_0 + 1 - i}$ and $A_{ii'} \sim \operatorname{N}(0, 1)$ for $i > i'$. Extracting all terms depending on $\Lambda^l$ in the ELBO (see Appendix \ref{appendixelbo}) and using the Bartlett decomposition, we get the following result:
\begin{prop}[Stochastic gate optimization]
\label{stochgradgateopt}
If $q^l(\Lambda) = \operatorname{W}(\Lambda \mid \Lambda^l, \eta_0)$, where $\Lambda^l = L^l L^{l\mathrm{T}}$ is the Cholesky factorization of $\Lambda^l$, updates are given by
\begin{equation*}
\begin{split}
L^{l+1} &= \; \underrel{\operatorname{arg\,min}}{L} \Bigg\{ -\eta_0 \log \det L \\
&\quad\quad + \sum_n \operatorname{\mathbb{E}}^{p(A)} \log\!\Bigg( \sum_{n' \neq n} \exp\!\Bigg( -\frac{1}{2}(x^n - x^{n'})^{\operatorname{T}} L A A^{\operatorname{T}} L^{\mathrm{T}} (x^n - x^{n'}) \Bigg) \Bigg) \\ 
&\quad\quad + \frac{\eta_0}{2} \operatorname{tr}\!\Bigg( L^{\operatorname{T}}\Bigg( \Lambda_0^{-1} + \sum_n \sum_{n' \neq n} \Omega_{nn'}^l (x^n - x^{n'})(x^n - x^{n'})^{\operatorname{T}} \Bigg) L \Bigg)  \Bigg\}.
\end{split}
\end{equation*}
\end{prop}
\noindent We will refer to the matrix in the parentheses between $L^{\operatorname{T}}$ and $L$ as the center matrix. Note that since $p(A)$ does not depend on $L$, a standard Monte Carlo approximation allows both the function value and its gradient to be estimated. This information may then be used in any gradient-based optimization algorithm to find the update $L^{l+1}$ and thus $\Lambda^{l+1}$, serving as an alternative to Proposition \ref{mstep2}. In this case, it is straightforward to derive an analogue to Proposition \ref{estep} by replacing the terms corresponding to the first transitions with their equivalents from the Monte Carlo estimation. Optionally, as evaluation of the involved quantities is relatively computationally expensive, we may reduce the variance of the standard Monte Carlo estimator by replacing it with an unbiased estimator with control variates introduced---for details, see Appendix \ref{control_variates}. Moreover, the center matrix, containing a double sum of outer vector products, may be evaluated in the alternative form displayed in Appendix \ref{center_matrix}.

\vspace{0.5cm}
\begin{algorithm2e}[H]
\label{algo}
\SetAlgoLined
\KwInput{} Training data $\{(x^n, y^n)\}_n$ \\
\KwOutput{} Variational posterior $q(\theta)$ \\
\KwInitialization{} $q^0(\{(u^n, z^n)\}_n)$, $q^0(\{(\mu_c, \Sigma_c)\}_c)$, $q^0(\Lambda)$, $s$, $t$ \\
\For{$l \in \{0, 1, \dots, l_{\max} - 1\}$}{
    \hangindent=1\skiptext\hangafter=1
    Optimize variational lower bound parameters $s^l$ according to (\ref{secondtransopt}) and, if not using stochastic gate optimization, $t^l$ according to (\ref{firsttransopt}) \\
    \emph{E-step}: Compute $q^{l+1}(\{(u^n, z^n)\}_n)$ from $q^l(\{(\mu_c, \Sigma_c)\}_c)$ and $q^l(\Lambda)$ according to Proposition \ref{estep} \\
    \emph{M-step I}: Compute $q^{l+1}(\{(\mu_c, \Sigma_c)\}_c)$ from $q^{l+1}(\{(u^n, z^n)\}_n)$ and $q^l(\Lambda)$ according to Proposition \ref{mstep1} \\
    \emph{M-step II}: Compute $q^{l+1}(\Lambda)$ from $q^{l+1}(\{(\mu_c, \Sigma_c)\}_c)$ and $q^{l+1}(\{(u^n, z^n)\}_n)$ according to Proposition \ref{mstep2} or \ref{stochgradgateopt} \\
}
$q(\theta) \leftarrow q^{l_{\max}}(\theta)$
\caption{Mean-field variational Bayes for the proposed model}
\end{algorithm2e}

\subsection{Posterior predictive distribution}

Having obtained the variational posterior approximation $q(\theta)$, prediction of the corresponding output $y^*$ given an out-of-sample input $x^*$ is determined by the posterior predictive distribution
\begin{equation*}
\begin{split}
p(y^* \mid x^*, \{(x^n, y^n)\}_{n}) &= \operatorname{\mathbb{E}}^{p(\theta \; \mid \; \{(x^n, y^n)\}_{n})} p(y^* \mid x^*, \{(x^n, y^n)\}_{n}, \theta) \\
&\approx \operatorname{\mathbb{E}}^{q(\theta)} p(y^* \mid x^*, \{(x^n, y^n)\}_{n}, \theta) \\
&= \sum_c \sum_{n} \operatorname{\mathbb{E}}^{q(\Lambda)} \frac{\operatorname{N}(x^* \mid x^n, \Lambda^{-1})}{\sum_{n'} \operatorname{N}(x^* \mid x^{n'}, \Lambda^{-1})} \\
&\quad\quad \times \operatorname{\mathbb{E}}^{q(\mu_c, \Sigma_c)} \frac{\operatorname{N}(y^n \mid \mu_c, \Sigma_c)}{\sum_{c'} \operatorname{N}(y^n \mid \mu_{c'}, \Sigma_{c'})}
\operatorname{N}(y^* \mid \mu_c, \Sigma_c).
\end{split}
\end{equation*}
As the marginalization integrals with respect to $q(\Lambda)$ and $q(\mu_c, \Sigma_c)$ are intractable, we will resort to Monte Carlo approximations. Note, in particular, that one may sample from $q(\Lambda)$ and each $q(\mu_c, \Sigma_c)$ independently due to the assumed factorization. Letting $\{\Lambda^k\}_{k=1}^{K_{\mathrm{g}}}$ and $\{(\mu_c^k, \Sigma_c^k)\}_{k=1}^{K_{\mathrm{e}}}$ be independent samples from $q(\Lambda)$ and each of the $q(\mu_c, \Sigma_c)$, respectively, we can write
\begin{align*}
p(y^* \mid x^*, \{(x^n, y^n)\}_{n})
&\approx \frac{1}{K_{\mathrm{e}} K_{\mathrm{g}}} \sum_{k=1}^{K_{\mathrm{e}}} \sum_{k'=1}^{K_{\mathrm{g}}} \sum_c \sum_n  \frac{\operatorname{N}(x^* \mid x^n, \Lambda^{-k'})}{\sum_{n'} \operatorname{N}(x^* \mid x^n, \Lambda^{-k'})} \\
&\quad\quad \times \frac{\operatorname{N}(y^n \mid \mu_c^k, \Sigma_c^k)}{\sum_{c'} \operatorname{N}(y^n \mid \mu_{c'}^k, \Sigma_{c'}^k)} \operatorname{N}(y^* \mid \mu_c^k, \Sigma_c^k).
\end{align*}
This results in a $K_{\mathrm{e}} C$-component Gaussian mixture as an approximate posterior predictive distribution.

\section{Computational study}
\label{seccomputationalstudy}

To illustrate the advantages of the proposed model, we evaluate its performance when applied on five datasets. The first two are artificially generated, intended to demonstrate the ability of the mixture-of-experts method to adapt to complex dependencies between the inputs and their corresponding predictive distributions and capture characteristics such as skewness and multimodality. The third is a dataset of dose statistics of radiation therapy treatment plans for postoperative prostate cancer, which is small, noisy and relatively high-dimensional, intended to showcase the stability of the estimated uncertainties even in such cases. The fourth dataset is the California housing dataset \citep{pace}, consisting of features for California housing districts along with geographic location---here, focusing on multimodal outputs, the task is to predict the latitude--longitude pairs for each district given its other features. The fifth dataset consists of infrared spectroscopy measurements for prediction of soil functional properties at different locations in Africa, aimed at illustrating the case of probabilistically predicting more than two output variables.

In the following computational study, attention was restricted to the stochastic gradient method described by Proposition \ref{stochgradgateopt} in Section \ref{stochgrad} for fitting the variational parameters associated with the gate, as the projection method described in Section \ref{stochgrad} was found to be insufficiently accurate. The method was implemented in Python---in particular, the stochastic gradient optimization used the automatic differentiation features and the Adam optimizer \citep{kingmaba} in Tensorflow 2.3.1. The hyperparameters of the priors were set and the variational parameters were initialized according to the default implementation outlined in Appendix \ref{hyperparameter_selection}---in particular, an agglomerative clustering algorithm \citep{sklearn} run on the standardized outputs $\{y^n\}_n$ was used to initialize the expert means. For all experiments, the maximum number of iterations in the mean-field variational Bayes algorithm was set to 20, each outer iteration comprising 50 stochastic gradient sub-iterations for the gate optimization. An early-stop mechanism was employed, terminating the algorithm if the inner optimization leaves the gate scale matrix non-significantly changed for three consecutive iterations---here, a non-significant change is defined as a run for which the inner iteration number and objective value has positive estimated correlation and for which non-correlation cannot be rejected at significance level $0.01$. All of the reported execution times are with respect to a computational service setup with a 12-core Intel Cascade Lake CPU platform, 85 GB of RAM and an NVIDIA A100 GPU with 40 GB of memory.

A comparison was made between the results obtained from the mixture-of-experts model and those from a conditional Gaussian DPMM model \citep{cruz-marzelo}, with posterior inference performed using a variational Bayes algorithm \citep{bleijordan, sklearn}, as well as a Gaussian process baseline model \citep{rasmussenwilliams, gpy}. For the DPMM, we used a Gaussian likelihood $p(y \mid \theta) = \operatorname{N}(y \mid \mu, \Sigma)$ with $\theta = (\mu, \Sigma)$, $\theta \mid G \sim G$ and $G \sim \operatorname{DP}(\alpha, G_0)$, where $\alpha = 1$ and $G_0$ was set to a conjugate normal--inverse-Wishart distribution, with hyperparameters selected analogously to the priors in our mixture-of-experts model. Using the stick-breaking representation of the Dirichlet process and a mean-field variational Bayes algorithm, the posterior predictive distribution may be Monte Carlo--approximated by a truncated Gaussian mixture \citep{bleijordan}. In particular, regarding each input--output concatenation $(x^n, y^n)$ in our dataset $\{(x^n, y^n)\}_n$ as a data point in the DPMM, the posterior predictive distribution $p(x^*, y^* \mid \{(x^n, y^n)\}_n)$ of a new pair $(x^*, y^*)$ is mixture-Gaussian, and the associated conditional distribution $p(y^* \mid x^*, \{(x^n, y^n)\}_n)$ is thus again mixture-Gaussian \citep{cruz-marzelo}. On the other hand, the Gaussian process baseline model used a radial basis function covariance $k : \mathcal{X}^2 \to \mathbb{R}$ with automatic relevance detection \citep{neal}, written as
\[
k(x, x') = \sigma^2 \operatorname{exp}\!\left( -\sum_i w_i (x_i - x'_i)^2 \right)
,\]
where the hyperparameters $\{w_i\}_i$ and $\sigma^2$ were fitted by evidence maximization. For the second and third datasets, where the output is multidimensional, Gaussian processes were fitted for the marginal distributions of the components separately, assuming independence between the components---this is done purely for simplicity, as the Gaussian process model mainly serves as a baseline method in our experiments. Apart from the KL divergence, the Hellinger distance $d_{\operatorname{H}}(p, q) = \Vert \sqrt{p} - \sqrt{q} \Vert_{L^2} / \sqrt{2}$ and the total variation distance $d_{\operatorname{TV}}(p, q) = \Vert p - q \Vert_{L^1} / 2$ were used as evaluation metrics for the test cases using synthetic data. The results of the comparison are shown in Table \ref{tableresults}.

\begin{table}[h]
\caption{Comparison of evaluated performance metrics between the proposed similarity-based mixture-of-experts (SBMoE) model, the conditional Dirichlet process mixture model (C-DPMM) and the baseline Gaussian process (GP) model. For the synthetic datasets, the KL divergence, the Hellinger distance and the total variation distance were used, whereas the mean negative log-likelihood was used for the other datasets due to the true predictive distribution being unknown.}
\centering
\begin{tabular}{llrrr}
\toprule
\multirow{2}{*}[-5pt]{Dataset} & \multirow{2}{*}[-5pt]{Evaluation metric} & \multicolumn{3}{c}{Model}\\
\addlinespace[0pt]
\cmidrule(lr){3-5} \\
\addlinespace[-10pt]
{} & {} & \multicolumn{1}{c}{SBMoE} & \multicolumn{1}{c}{C-DPMM} & \multicolumn{1}{c}{GP} \\
\midrule
\multirow{3}{*}{1D synthetic} & KL divergence ($10^{-2}$) & $0.96$ & $5.64$ & $39.12$ \\
& Hellinger distance ($10^{-2}$) & $4.62$ & $11.80$ & $24.80$ \\
& Total variation ($10^{-2}$) & $1.51$ & $3.95$ & $8.12$ \\ 
\cmidrule(lr){1-5}
\multirow{3}{*}{2D synthetic} & KL divergence ($10^{-3}$) & $2.43$ & $19.08$ & $3.27$  \\
& Hellinger distance ($10^{-2}$) & $2.33$ & $4.77$ & $2.52$ \\
& Total variation ($10^{-3}$) & $1.71$ & $3.73$ & $1.86$ \\ 
\cmidrule(lr){1-5}
\multirow{1}{*}{Radiation therapy} & Mean negative log-likelihood & $-2.24$ & $-2.29$ & $-1.66$ \\
\cmidrule(lr){1-5}
\multirow{1}{*}{California housing} & Mean negative log-likelihood & $-0.34$ & $0.46$ & $2.66$ \\
\cmidrule(lr){1-5}
\multirow{1}{*}{Africa soil} & Mean negative log-likelihood & $4.29$ & $72.79$ & $7.43$ \\
\bottomrule
\end{tabular}
\label{tableresults}
\end{table}

\subsection{One-dimensional synthetic data}
\label{1dtoydata}

\begin{figure}[h]
\centering
\includegraphics[width=\textwidth]{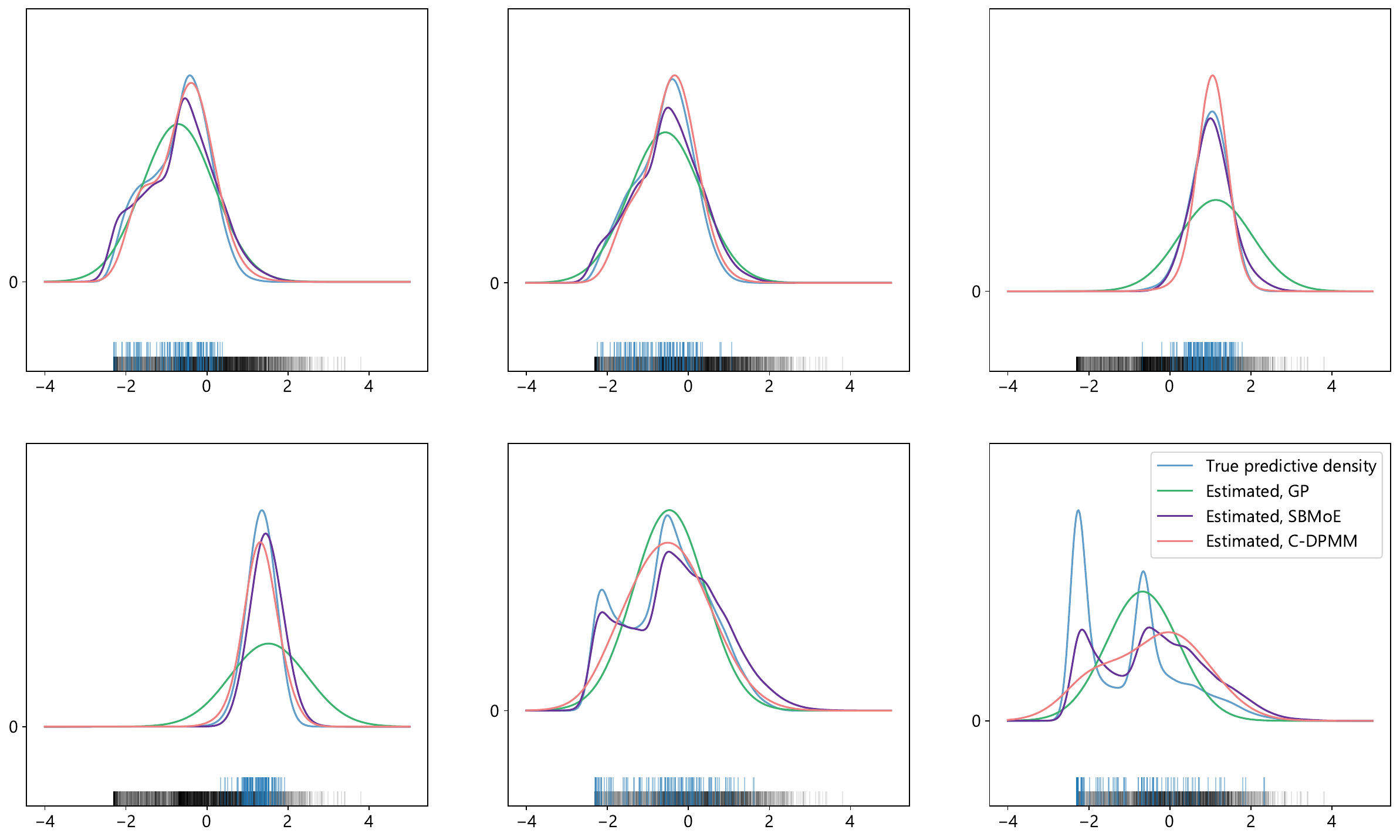}
\caption{Comparison between a kernel density estimate of the true predictive density $p(y^* \mid x^*, \{(x^n, y^n)\}_n)$ and the estimated counterparts from the mixture-of-experts model, the conditional DPMM and the Gaussian process, shown for six out-of-sample inputs $x^*$ of the one-dimensional synthetic data. The gray bars below the graph represent the training data points $\{y^n\}_n$, and the blue bars represent samples from the true predictive distribution.}
\label{pred_1d_1}
\end{figure}

In this example, the observed data $\{(x^n, y^n)\}_{n=1}^N$ consists of $N = 2000$ exchangeable copies of $(x, y) = ((x_1, x_2), y)$ following the generative model 
\begin{align*}
x &\sim \operatorname{LN}\!\left(
\begin{bmatrix}
0 \\
0
\end{bmatrix},
\begin{bmatrix}
1 & 0.5 \\
0.5 & 1
\end{bmatrix}
\right), \\
\tau &\sim \operatorname{Be}(0.3), \\
\zeta \mid x &\sim \Gamma(x_1, x_2), \\
y &= \log(\zeta + 0.4\tau + 0.1),
\end{align*}
with $\tau$ independent of $x$ and $\zeta$, where we use $\operatorname{LN}$, $\operatorname{Be}$ and $\Gamma$ to denote the log-normal, the Bernoulli and the gamma distribution (with shape--rate parameterization), respectively. As such, the predictive distributions are both skewed and possibly bimodal. The intent of choosing a relatively large data size was to demonstrate the capability of the proposed mixture-of-experts model to approach the complex predictive distributions with growing data---for smaller datasets, while the methods still work well, we found it hard to evaluate and compare the methods in any meaningful way. Using $C = 32$ experts for our proposed mixture-of-experts model (note that the expert locations are fixed with respect to $x$, thus the need for a relatively high number), approximate posteriors were computed using the proposed mean-field variational Bayes algorithm. The training of the mixture-of-experts, Gaussian process and conditional DPMM models took 553, 14 and 75 seconds on the aforementioned computational setup. A comparison of the estimated predictive distributions is shown in Figure \ref{pred_1d_1} alongside a kernel density estimate of the true predictive density using 5000 samples from the true predictive distribution, a Gaussian kernel and Scott's rule \citep{scott} for bandwidth selection. The evaluated performance metrics are summarized in Table \ref{tableresults}. It is apparent that our mixture-of-experts model performed better than the conditional DPMM---which was, in turn, better than the Gaussian process model---in terms of performance metrics as well as visual inspection. Specifically, the flexibility of our model allows for much better fit to the bimodal and skewed distributions even compared to the conditional DPMM, which was often not able to capture bimodal behavior despite also having mixture-Gaussian predictive distributions.

\subsection{Two-dimensional synthetic data}
\label{2dtoydata}

\begin{figure}[h]
\centering
\includegraphics[width=\textwidth]{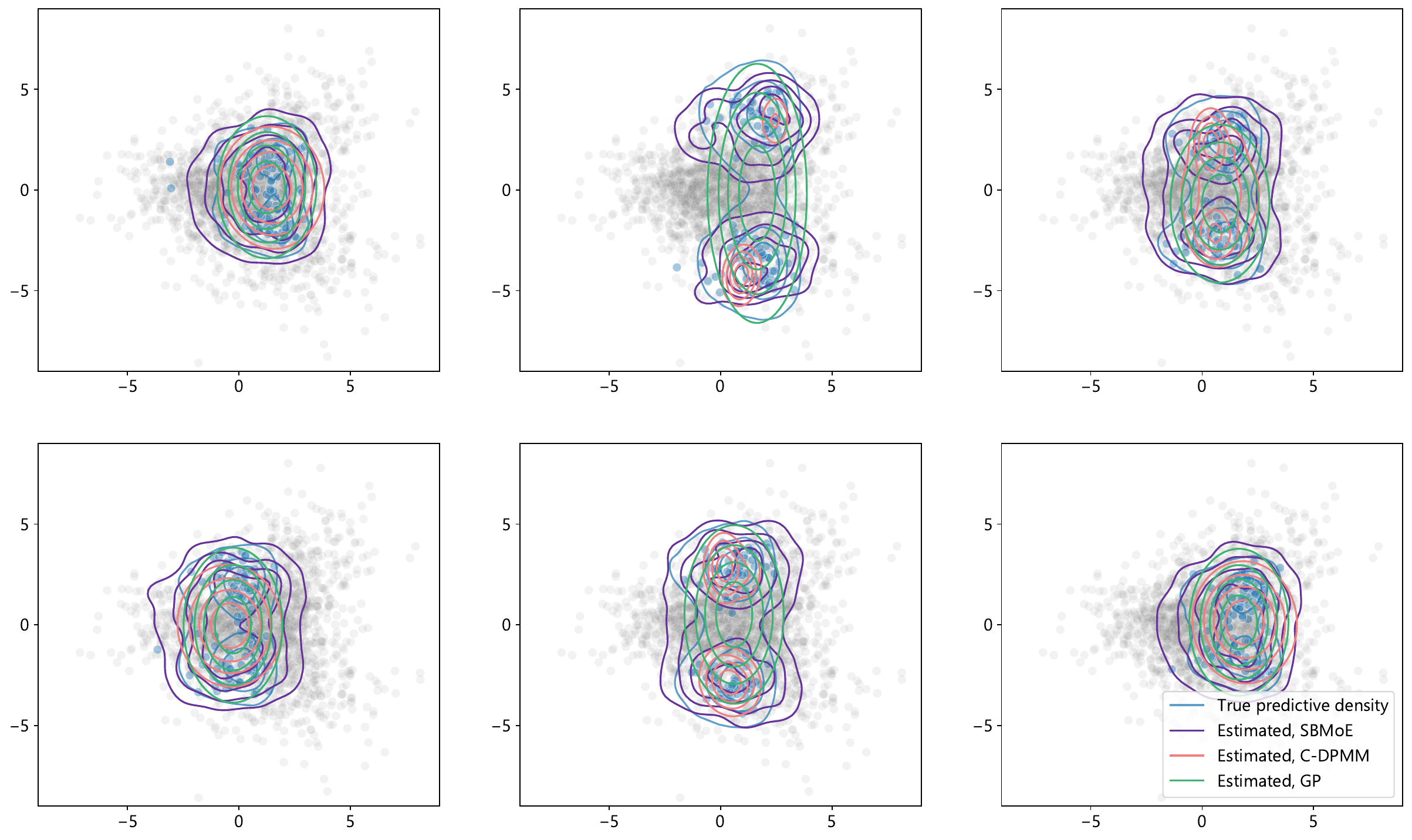}
\caption{Comparison between a kernel density estimate of the true predictive density $p(y^* \mid x^*, \{(x^n, y^n)\}_n)$ and the estimated counterparts from the mixture-of-experts model, the conditional DPMM and the Gaussian process, shown for six out-of-sample inputs $x^*$ of the two-dimensional synthetic data. The gray dots represent the training data points $\{y^n\}_n$, and the blue dots represent samples from the true predictive distribution.}
\label{pred_2d_1}
\end{figure}

To illustrate the advantages of our model in estimating predictive distributions with inter-component dependencies, we now use a $32$-dimensional input $x = (x_i)_{i=1}^{32}$ and a two-dimensional output $y = (y_1, y_2)$. Define the transformation $\varphi(\xi) = (\operatorname{N}(0.7 i \mid \xi, 4))_{i=0}^7$, and denote by $\mathcal{F}$ the discrete Fourier transform. The observed data consists of $N = 2000$ data points drawn from the generative model
\begin{equation*}
\begin{split}
\eta = (\eta_1, \eta_2) &\sim \operatorname{LN}\!\left(
\begin{bmatrix}
0 \\
0
\end{bmatrix},
\begin{bmatrix}
1 & 0.5 \\
0.5 & 1
\end{bmatrix}
\right), \\
x &= \left( \operatorname{Re} \mathcal{F}(\varphi(\eta_1)), \operatorname{Im} \mathcal{F} (\varphi(\eta_1)), \operatorname{Re} \mathcal{F} (\varphi(\eta_2)), \operatorname{Im} \mathcal{F} (\varphi(\eta_2)) \right), \\
\tau &\sim \operatorname{Be}(0.5), \\
\zeta = (\zeta_{ii'})_{i, i'=1}^2 \mid \eta &\sim \operatorname{IW}\!\left( 
\begin{bmatrix}
\eta_1 & 0 \\
0.5 & \eta_2
\end{bmatrix}
\begin{bmatrix}
\eta_1 & 0 \\
0.5 & \eta_2
\end{bmatrix}^{\operatorname{T}}\!,
3
\right), \\
y &= (\log \zeta_{11}, (2\tau - 1) \log \zeta_{22}),
\end{split}
\end{equation*}
where $\tau$ is independent of $\eta$ and $\zeta$. In particular, the log-normal generator $\eta$ is observed through its obsfucated counterpart $x$---the map $\varphi$ creates spikes centered at its inputs $\eta_1$, $\eta_2$ represented as eight-dimensional vectors, whose real and imaginary Fourier components are concatenated into a $32$-dimensional $x$. Given $\eta$, we then draw $\zeta$ as an inverse-Wishart matrix whose diagonal components are log-transformed and put into $y$, with the sign of the second component flipped with probability $0.5$. Again, although all methods remained stable for smaller datasets, a relatively large data size was found necessary for the comparison to clearly showcase key qualities of the methods. Using $C = 64$ classes and running the mean-field variational Bayes algorithm, we obtain the results visualized in Figure \ref{pred_2d_1}, where estimated densities for our mixture-of-experts model, the conditional DPMM and the per-component Gaussian process model are shown in comparison to a kernel density estimate of the true predictive density, using 10000 samples from the true predictive distribution and again a Gaussian kernel and Scott's rule for bandwidth selection. The training of the mixture-of-experts, Gaussian process and conditional DPMM models took 1116, 311 and 245 seconds. The evaluated performance metrics are summarized in Table \ref{tableresults}. Again, besides from having better validation metrics, our model is clearly more successful than the Gaussian process model in capturing the characteristics of the bimodal shape of the true predictive distribution. More surprisingly, the conditional DPMM, while also being able to recognize bimodal distributions, yielded much worse performance metrics than both our and the Gaussian process model. This may be explained by the conditional DPMM's generative modeling using mixture models, which presents difficulties in settings in which inputs are of relatively high dimension but sparsely distributed---note that each $32$-dimensional $x$ is uniquely determined by the associated two-dimensional $\eta$. In turn, this may lead to the underestimation of uncertainty we observe for the conditional DPMM, causing the considerably worse performance metrics. In contrast, the mixture-of-experts and Gaussian process models are both discriminative and similarity-based and therefore relatively insensitive to the complexities of the input space. 

\subsection{Radiation therapy data}

\begin{figure}[h]
\centering
\includegraphics[width=\textwidth]{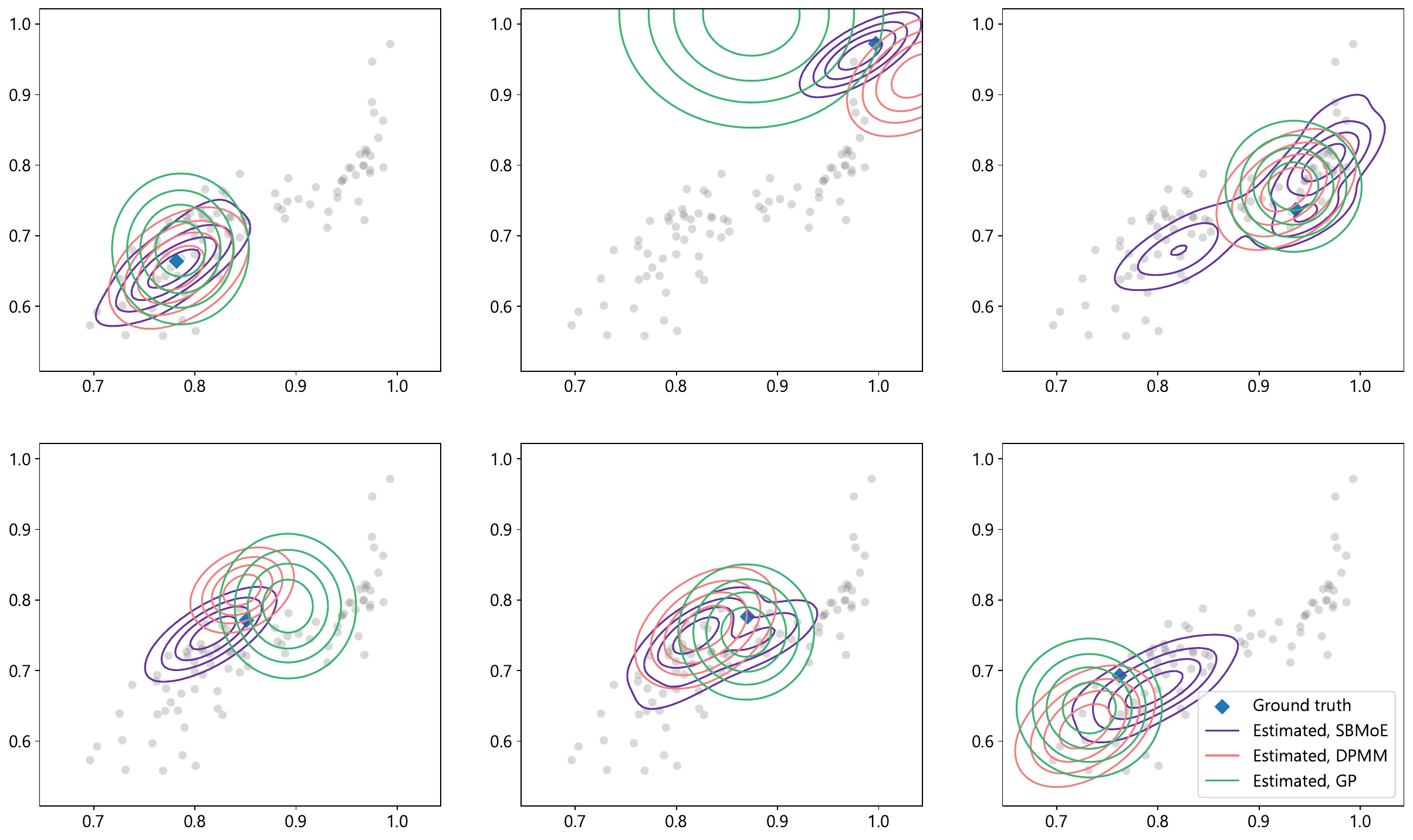}
\caption{Comparison between the estimated predictive densities $p(y^* \mid x^*, \{(x^n, y^n)\}_n)$ from the mixture-of-experts model, the conditional DPMM and the Gaussian process, shown for six random out-of-sample inputs $x^*$ of the radiation therapy data. The gray dots represent the training data points $\{y^n\}_n$, and the blue diamond shows the ground truth $y^*$. Axes represent fraction of the prescription dose.}
\label{pred_dvh_1}
\end{figure}

\begin{figure}[h]
\centering
\includegraphics[width=\textwidth]{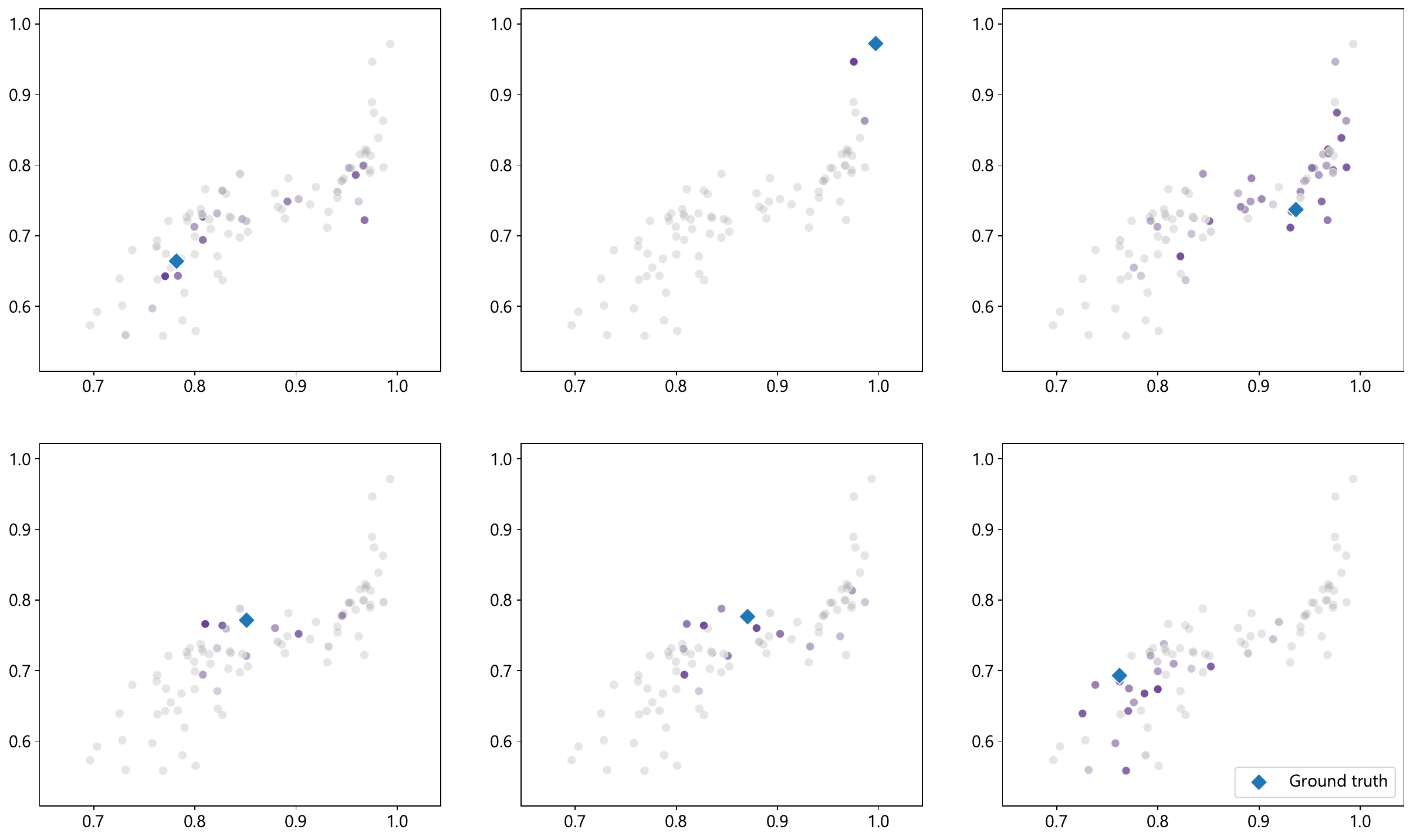}
\caption{Visualization of the computed distances $d_{\mathcal{X}}(x^*, x^n)$ to each training input $x^n$, shown for the same out-of-sample inputs $x^*$ of the radiation therapy data as in Figure \ref{pred_dvh_1}. A darker purple color corresponds to a smaller distance. Axes represent fraction of the prescription dose.}
\label{pred_dvh_2}
\end{figure}

We also demonstrate our method on a dataset of historically delivered radiation therapy treatment plans for postoperative prostate cancer patients from the Iridium Cancer Network in Antwerp, Belgium. Prediction of radiation dose statistics based on patient geometry is of particular interest in fields such as automated treatment planning and quality assurance \citep{ge}. The dataset consists of 94 treatment plans on which features have been extracted from patient images and dose statistics have been computed for the treatment plans, of which 84 were used as training data. For our purposes, attention was restricted to the bladder region---the features we used are signed distance transforms computed from the bladder region to itself, the prostate planning target volume, the seminal vesicles planning target volume and the rectum region, binned into a total of 44 features; the dose statistics we used are the dose-at-volume values at $40$ and $20$ percent in the bladder region, corresponding to the $0.6$- and $0.8$-level quantiles of the dose delivered to the region. Using $C = 32$ mixture classes, for the mixture-of-experts, Gaussian process and conditional DPMM models, the training took 393, 2 and 1 seconds, respectively. The results of the comparison to the conditional DPMM and the Gaussian process model are presented in Table \ref{tableresults} and Figures \ref{pred_dvh_1} and \ref{pred_dvh_2}. In terms of mean negative log-likelihood over the test dataset, the mixture-of-experts model had comparable and only slightly better performance to the conditional DPMM the Gaussian process model, respectively. More importantly, however, it managed to follow the S-shaped distribution of the data points in its predictions, whereas the Gaussian process model sometimes predicted far outside the range of reasonable values. From Figure \ref{pred_dvh_2}, we also see that the learned distance metric $d_{\mathcal{X}}$ produced a varying number of near neighbors, implying an accordingly varying degree of uncertainty. It is especially interesting to note that this was possible using such a small training dataset with relatively high input dimensions, which is seen as a sign of robustness against data scarcity.

\subsection{California housing data}

\begin{figure}[h]
\centering
\includegraphics[width=\textwidth]{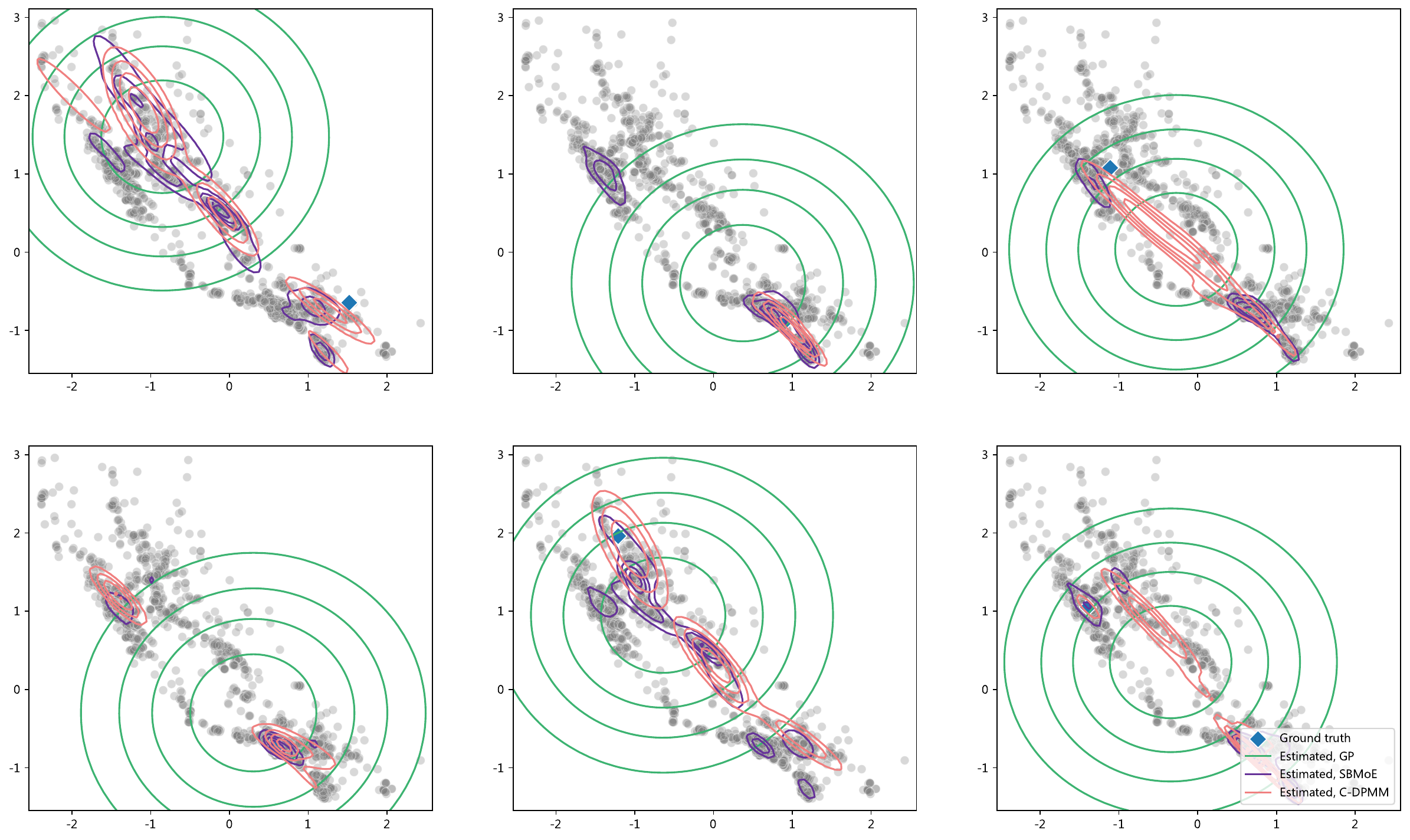}
\caption{Predictive distributions of geographic location for the SBMoE, C-DPMM and GP models versus ground truth for the California housing data.}
\label{california}
\end{figure}

Here, we consider the California housing price dataset \citep{pace}, which comprises variables for 20640 housing districts in California, United States. In particular, for each district, we have the longitude--latitude pair of its geographical location, the median age, the total number of rooms, the total number of bedrooms, the population, the number of households, the median income, the median house value and the ocean proximity. Removing irregular data points, the categorical ocean proximity variable and replacing the total number of rooms and bedrooms by the corresponding mean per household, 18299 data points remain, with seven input and two output dimensions. Furthermore, subsampling with uniform probability, we use 3000 training and 100 test data points for the numerical experiments. For our purposes, we will try to predict the latitude--longitude pair of each out-of-sample district given the other seven features. This is a non-straightforward task for several reasons: the signal-to-noise ratio is relatively small with the available input features, the map from features to latitude--longitude pair is likely rather complex, and the predictive distributions are in general multimodal as similar housing districts may occur in areas far away from each other. Using $C = 32$, training of the mixture-of-experts model took 194 seconds, compared to 196 seconds for the Gaussian process and 229 seconds for the conditional DPMM. Table \ref{tableresults} contains performance metrics for the different methods, and Figure \ref{california} shows example predictions for six random out-of-sample test inputs. We see that the mixture-of-experts model and the conditional DPMM, in contrast to the Gaussian process, are able to produce multimodal predictive distributions following the geographical shape of the data, successfully identifying housing district clusters such as cities. Moreover, the mixture-of-experts model performs better in terms of mean negative log-likelihood than the other models. In particular, as is the case for the two-dimensional synthetic dataset, the conditional DPMM tends to slightly underestimate the predictive uncertainties.

\subsection{Africa soil data}

\begin{figure}[h]
\centering
\includegraphics[width=\textwidth]{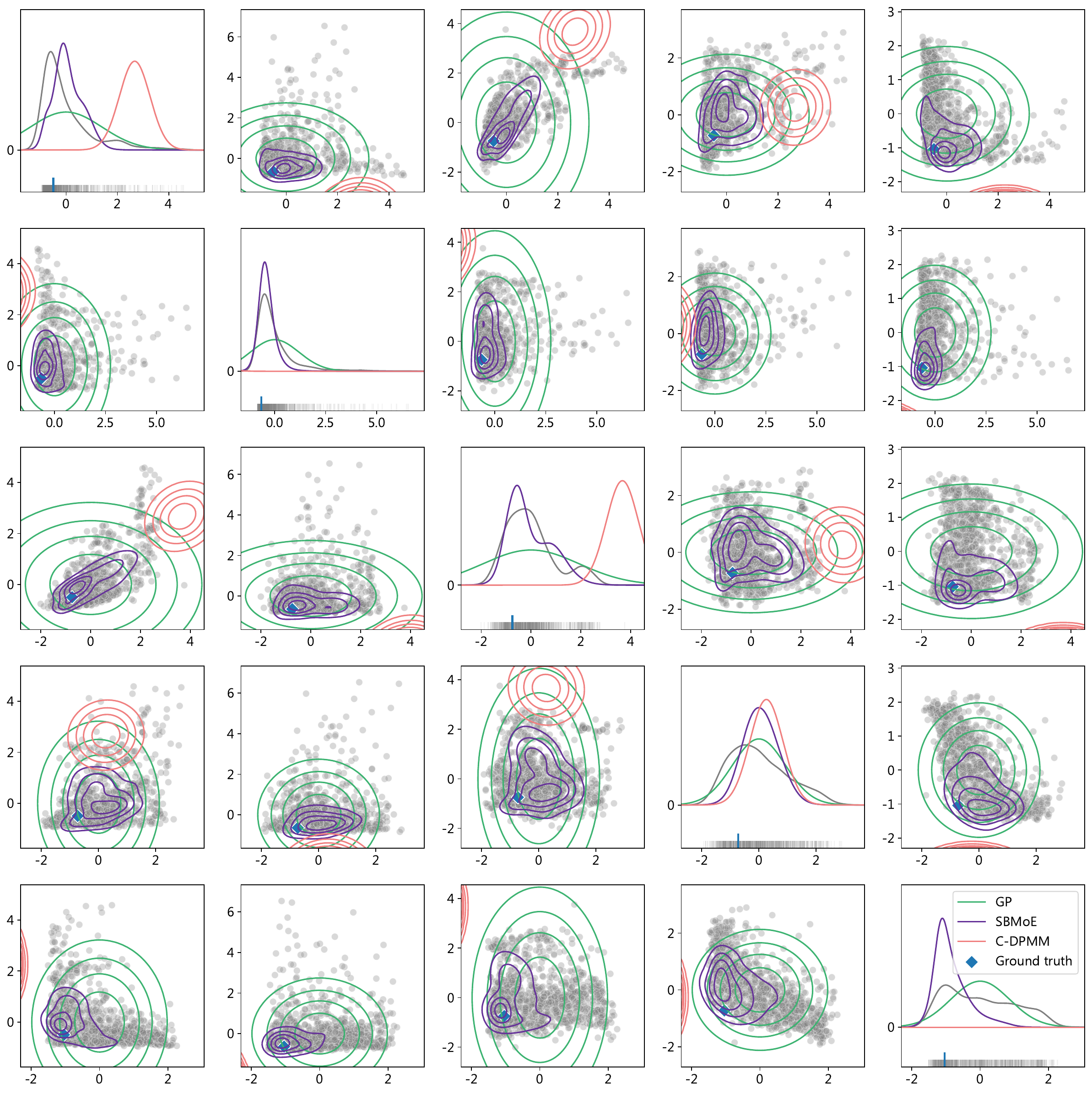}
\caption{Predictive distributions for the SBMoE, C-DPMM and GP models versus ground truth for the Africa soil data.}
\label{africa_soil_multidim}
\end{figure}

In this last experiment, we consider the prediction of soil functional properties from infrared spectroscopy measurements as a means of rapid, low-cost analysis of soil samples. Especially in data-sparse regions such as Africa, such tools are important for planning sustainable agricultural intensification and natural resource management. The dataset \citep{africa_soil} contains 1157 data points, each having 3578 dimensions of diffuse-reflectance infrared spectroscopy measurements at different wavelengths, 16 features related to the geography of the location at which the sample is collected, and five output variables comprising the soil organic carbon, the pH value and the calcium, phosphorous and sand content---for more details, see \citet{africa_soil}. Using 1000 data points for training and excluding the rest for testing, the 3578-dimensional spectroscopy measurements are dimensionality-reduced by a multi-level discrete wavelet transform \citep{mallat} using a Daubechies 4 wavelet into 74-dimensional representations---thus, the problem considered amounts to having $\operatorname{dim} \mathcal{X} = 90$ and $\operatorname{dim} \mathcal{Y} = 5$. With $C = 64$ classes, the trainings for the mixture-of-experts model, the Gaussian process and the conditional DPMM ran for 803, 1279 and 24 seconds, respectively. Table \ref{tableresults} shows the evaluated performance metrics, and Figure \ref{africa_soil_multidim} shows the single and pairwise marginal predictive distributions for a randomly chosen test input. Again, the mixture-of-experts model performs better than the competitor models, with the conditional DPMM obtaining a particularly bad score. We can see from the example prediction in Figure \ref{africa_soil_multidim} that our mixture-of-experts model is able to follow the data distribution well and predict the ground truth accurately with varying degrees of uncertainty for the different output dimensions---in comparison, the Gaussian process is only able to output an ill-fitted Gaussian predictive distribution, and the predictions from the conditional DPMM are mostly far away from the ground truth. Indeed, it is reasonable to believe that the particularly bad performance of the conditional DPMM is due to the relatively high dimension of the data, where fitting a mixture model to $\operatorname{dim} \mathcal{X} + \operatorname{dim} \mathcal{Y} = 95$ dimensions and taking the conditional given a 90-dimensional input is likely to be extremely sensitive to calibration errors. 

\section{Discussion}
\label{secdiscussion}

Motivated by the disadvantages of commonly used parametric models when applied to data with complex input--output relationships, in this paper, we have built upon previous work on probabilistic $k$-nearest neighbors by \citet{holmes}, \citet{cucala} and \citet{friel} by extending the ideas to a regression setting, using Gaussian mixtures as a flexible means of representing predictive distributions. We use a conditionally specified model in which full conditionals are defined and the joint likelihood is replaced by a pseudolikelihood in order to render further computations tractable. In particular, in contrast to \citet{friel}, we also regard the precision matrix $\Lambda$ in the distance metric $d_{\mathcal{X}}$ as a model parameter on which posterior inference may be performed. With a nonparametric approach, we avoid the need to explicitly learn the mapping between input and output as, for example, in conventional mixture-of-experts models such as those described in \citet{jacobsjordan}, \citet{bishopsvensen} and \citet{xu}, all while maintaining a Bayesian approach. Also in contrast to, e.g., the probabilistic $k$-nearest neighbor models by \citet{holmes}, \citet{cucala} and \citet{friel} as well as the DPMM-based models by, e.g., \citet{deiorio_linear}, \citet{dunson}, \citet{jara} and \citet{cruz-marzelo}, we exploit the mixture-Gaussian structure of our model and base our parameter inference on a mean-field variational Bayes algorithm, where local variational approximations are introduced and associated variational parameters are optimized by linear programming and stochastic gradient descent methods. Thus, we avoid the need for input-to-parameter modeling present in DPMM-based models, which in our case is a multivariate regression problem from dimension $\operatorname{dim} \mathcal{X}$ to $\operatorname{dim }\mathcal{Y} + \operatorname{dim} \mathcal{Y}^2$---in principle, potentially even harder than the original problem from dimension $\operatorname{dim} \mathcal{X}$ to $\mathcal{Y}$. Furthermore, we avoid the use of MCMC methods, which scale poorly with model size and data dimensionality. In contrast to the conditional DPMM, which models the distribution of the inputs generatively by a mixture model on $\mathcal{X} \oplus \mathcal{Y}$, our method relies on discriminative modeling by comparing similarities between inputs. The computational study demonstrates on several synthetic as well as real-world datasets the ability to model irregular, multivariate predictive densities, while being able to quantify uncertainties accurately for medium-sized to very small datasets. Specifically, we showed that our mixture-of-experts model outperforms the conditional DPMM and the Gaussian process baseline in certain settings---for example, those in which the input space is relatively high-dimensional yet sparse and thus unsuitable for generative modeling using mixture models. While generally being more computationally demanding than the conditional DPMM and Gaussian process, the execution times for training the mixture-of-experts model on the given computational setup were certainly short for a high-dimensional Bayesian model.

Based on the method we describe in this paper, there are ample opportunities for interesting future work. A drawback of the method is its lack of scalability to massive data, an issue common for all nonparametric models where the training data is explicitly used in the predictive pipeline. In our case, the main difficulties stem from the need to compute the posterior probabilities $\omega_{c, nn'}^l$ at each iteration $l$, which requires storage in an array of $C N^2$ elements. One possible approach to addressing this is to use Bayesian coresets \citep{huggins}, which is a small weighted subset of the data constructed so as to optimally summarize the data; another is to take inspiration from the sparse pseudo-input Gaussian process literature \citep{quinonerocandela}. A further drawback is the subtle but important assumption that the input space $\mathcal{X}$ is a vector space---while this may be true for all datasets considered in our computational study, when handling e.g. image data, one would need some form of preprocessing by a feature-extracting transformation $\phi : \mathcal{I} \to \mathcal{X}$ from the image space $\mathcal{I}$ to the vector space $\mathcal{X}$. If, for example, $\phi$ is a neural network, one can optimize the network weights using the same optimization problem as in Proposition \ref{stochgradgateopt} by regarding the weights as hyperparameters. One may also investigate sparse representations of the scale matrix $\Lambda^l$ for the case of $\operatorname{dim} \mathcal{X}$ being too large to store the full Cholesky factorization of $\Lambda^l$. Lastly, to address the problem of choosing an appropriate number of experts, yet another direction of future work would be to explore the possibility of combining the present model with Dirichlet process mixtures, further conforming to the philosophy of Bayesian nonparametrics.

\begin{appendices}

\section{Variational posteriors}
\label{secappendix}

We start by recalling some facts which will be useful for the derivations. The densities of the Wishart and inverse-Wishart distributions may be written as
\[
\operatorname{W}(X \mid X_0, \nu_0) = \frac{1}{2^{\nu_0 d / 2} (\det X_0)^{\nu_0 / 2} \Gamma_{d}\!\left( \frac{\nu_0}{2} \right)} (\det X)^{(\nu_0 - d - 1) / 2} e^{ -\operatorname{tr}(X_0^{-1} X) / 2 }
\]
and
\[
\operatorname{IW}(X \mid X_0, \nu_0) = \frac{(\det X_0)^{\nu_0 / 2}}{2^{\nu_0 d / 2} \Gamma_{d}\!\left( \frac{\nu_0}{2} \right)} (\det X)^{-(\nu_0 + d + 1) / 2} e^{ -\operatorname{tr}(X_0 X^{-1}) / 2 }
,\]
where $X$ and $X_0$ are positive definite $d \times d$ matrices, $\nu_0 > d - 1$ is a constant and $\Gamma_d$ is the multivariate gamma function \citep{anderson}. In particular, $X \sim \operatorname{W}(X_0, \nu_0)$ if and only if $X^{-1} \sim \operatorname{IW}(X_0^{-1}, \nu_0)$. We will frequently use the following results \citep{anderson, bishop}:
\begin{lemm}
\label{thethree}
If $\Sigma \sim \operatorname{IW}(\Sigma_0, \nu_0)$ is a $d \times d$ matrix and $\mu \mid \Sigma \sim \operatorname{N}(\mu_0, \kappa_0^{-1} \Sigma)$, then
\begin{enumerate}[label=(\roman*)]
\item $\operatorname{\mathbb{E}} \Sigma^{-1} = \nu_0 \Sigma_0^{-1}$,
\item $\operatorname{\mathbb{E}} \log \det \Sigma = -\displaystyle\sum_{i=1}^{d} \psi\!\left( \frac{\nu_0 + 1 - i}{2} \right) - d \log 2 + \log \det \Sigma_0$, where $\psi$ is the digamma function, and
\item $\operatorname{\mathbb{E}} \, (y - \mu)^{\operatorname{T}} \Sigma^{-1}(y - \mu) = \dfrac{d}{\kappa_0}  + \nu_0 (y - \mu_0)^{\operatorname{T}} \Sigma_0^{-1} (y - \mu_0)$ for all vectors $y$. 
\end{enumerate}
\end{lemm}

\subsection{Proof of Proposition \ref{estep}}
\label{proofestep1}

Starting from the log-joint in (\ref{logjoint}), we apply the local variational approximations (\ref{secondtransapprox}), (\ref{firsttransapprox}) and, at iteration $l$, set $s$ and $t$ to optimized values $s^l$ and $t^l$ to obtain an approximate log-pseudojoint. Disregarding terms not depending on $\{(u^n, z^n)\}_n$, we have
\begin{align*}
\log & \, q^{l+1}(\{(u^n, z^n)\}_n) \\
&\stackrel{\mathrm{c}}{=} \operatorname{\mathbb{E}}^{q^l(\Lambda) q^l(\{(\mu_c, \Sigma_c)\}_c)} \log p(\{(y^n, u^n, z^n)\}_n, \theta \mid \{x^n\}_n) \\
&\stackrel{\mathrm{c}}{\approx} \sum_n \sum_{n' \neq n} \sum_{c} 1_{z^n \; = \; c} 1_{u^n \; = \; n'} \operatorname{\mathbb{E}}^{q^l(\Lambda) q^l(\{(\mu_c, \Sigma_c)\}_c)}\!\Bigg[ \log \operatorname{N}(y^n \mid \mu_c, \Sigma_c) \\
&\quad\quad + \log \operatorname{N}(y^{n'} \mid \mu_c, \Sigma_c) - \sum_{c'} s_{n'c'}^l \log \operatorname{N}(y^{n'} \mid \mu_{c'}, \Sigma_{c'}) \\
&\quad\quad + \log \operatorname{N}(x^n \mid x^{n'}, \Lambda^{-1}) - \sum_{n'' \neq n} t_{nn''}^l \log \operatorname{N}(x^n \mid x^{n''}, \Lambda^{-1}) \Bigg] \\
&\stackrel{\mathrm{c}}{=} \sum_n \sum_{n' \neq n} \sum_{c} 1_{z^n \; = \; c} 1_{u^n \; = \; n'} \Bigg( \\
&\quad\quad -\operatorname{\mathbb{E}}^{q^l(\Sigma_c)} \log \det \Sigma_c + \frac{1}{2} \sum_{c'} s_{n'c'}^l \operatorname{\mathbb{E}}^{q^l(\Sigma_{c'})} \log \det \Sigma_{c'} \\
&\quad\quad -\frac{1}{2} \operatorname{\mathbb{E}}^{q^l(\mu_c, \Sigma_c)}\!\left[ (y^n - \mu_c)^{\operatorname{T}} \Sigma_c^{-1} (y^n - \mu_c) + (y^{n'} - \mu_c)^{\operatorname{T}} \Sigma_c^{-1} (y^{n'} - \mu_c) \right] \\
&\quad\quad +\frac{1}{2} \sum_{c'} s_{n'c'}^l \operatorname{\mathbb{E}}^{q^l(\mu_{c'}, \Sigma_{c'})} (y^{n'} - \mu_{c'})^{\operatorname{T}} \Sigma_{c'}^{-1} (y^{n'} - \mu_{c'}) \\
&\quad\quad -\frac{1}{2} \operatorname{\mathbb{E}}^{q^l(\Lambda)} (x^n - x^{n'})^{\operatorname{T}} \Lambda (x^n - x^{n'}) \\
&\quad\quad +\frac{1}{2} \operatorname{\mathbb{E}}^{q^l(\Lambda)} \sum_{n'' \neq n} t_{nn''}^l (x^n - x^{n''})^{\operatorname{T}}  \Lambda (x^n - x^{n''}) \Bigg),
\end{align*}
which is identified as the probability mass function of a (three-dimensional) categorical distribution---note, in particular, that this conclusion may be drawn regardless of the forms of the distributions $q^l(\Lambda)$ and $q^l(\{(\mu_c, \Sigma_c)\}_c)$. Computing the expectations according to Lemma \ref{thethree}, the expressions for $\log \omega_{c, nn'}^{l+1}$ follow. 

\subsection{Proof of Proposition \ref{mstep1}}
\label{proofmstep1}

Again, starting from the log-pseudojoint in (\ref{logjoint}), we apply the local variational approximations (\ref{secondtransapprox}), (\ref{firsttransapprox}), use the fact that $\operatorname{\mathbb{E}}^{q^l(u^n, z^n)} 1_{z^n \; = \; c} 1_{u^n \; = \; n'} = \omega_{c, nn'}^l$, collect relevant terms and exchange orders of summation to obtain
\begin{align*}
\log & \, q^{l+1}(\{(\mu_c, \Sigma_c)\}_c) \\
&\stackrel{\mathrm{c}}{=} \operatorname{\mathbb{E}}^{q^{l+1}(\{(u^n, z^n)\}_n) q^l(\Lambda)} \log p(\{(y^n, u^n, z^n)\}_n, \theta \mid \{x^n\}_n) \\
&\stackrel{\mathrm{c}}{\approx} \sum_c \left( \log \operatorname{N}\!\left(\mu_c \; \bigg| \; \mu_0, \frac{1}{\kappa_0} \Sigma_c\right) + \log \operatorname{IW}(\Sigma_c \mid \Sigma_0, \nu_0) \right) \\
&\quad\quad + \sum_n \sum_{n' \neq n} \sum_{c} \omega_{c, nn'}^{l+1} \Bigg( \log \operatorname{N}(y^n \mid \mu_c, \Sigma_c) + \log \operatorname{N}(y^{n'} \mid \mu_c, \Sigma_c)  \\
&\quad\quad\quad\quad - \sum_{c'} s_{n'c'}^{l+1} \log \operatorname{N}(y^{n'} \mid \mu_{c'}, \Sigma_{c'}) \Bigg) \\
&\stackrel{\mathrm{c}}{=} \sum_c \Bigg( \bigg( -\frac{\nu_0 + \dim \mathcal{Y} + 2}{2} - \sum_n \sum_{n' \neq n} \bigg( \omega_{c, nn'}^{l+1} - \frac{1}{2} s_{n'c}^{l+1} \Omega_{nn'}^{l+1} \bigg) \bigg) \log \det \Sigma_c \\
&\quad\quad -\frac{1}{2} \operatorname{tr}(\Sigma_0 \Sigma_c^{-1}) - \frac{\kappa_0}{2} (\mu_c - \mu_0)^{\operatorname{T}} \Sigma_c^{-1} (\mu_c - \mu_0) \\
&\quad\quad - \frac{1}{2} \sum_n \sum_{n' \neq n} \bigg( \omega_{c, nn'}^{l+1} (y^n - \mu_c)^{\operatorname{T}} \Sigma_c^{-1} (y^n - \mu_c) \\
&\quad\quad\quad\quad + \omega_{c, nn'}^{l+1} (y^{n'} - \mu_c)^{\operatorname{T}} \Sigma_c^{-1} (y^{n'} - \mu_c)  \\
&\quad\quad\quad\quad - s_{n'c}^{l+1} \Omega_{nn'}^{l+1} (y^{n'} - \mu_c)^{\operatorname{T}} \Sigma_c^{-1} (y^{n'} - \mu_c) \bigg) \Bigg) \\
&= \sum_c \Bigg( -\frac{\nu_0 + \dim \mathcal{Y} + 2 + R_c^{l+1}}{2} \log \det \Sigma_c - \frac{1}{2} \operatorname{tr}(\Sigma_0 \Sigma_c^{-1}) \\
&\quad\quad - \frac{\kappa_0}{2} (\mu_c - \mu_0)^{\operatorname{T}} \Sigma_c^{-1} (\mu_c - \mu_0)  - \frac{1}{2} \sum_n r_{nc}^{l+1} (y^n - \mu_c)^{\operatorname{T}} \Sigma_c^{-1} (y^n - \mu_c) \Bigg).
\end{align*}
Note that this leads to the sought decomposition $q^l(\{(\mu_c, \Sigma_c)\}_c) = \prod_c q^l(\mu_c \mid \Sigma_c) q^l(\Sigma_c)$ for all $l$. In particular, this implies that
\[
\log q^l(\mu_c \mid \Sigma_c) \stackrel{\mathrm{c}}{=} -\frac{1}{2} \left( \kappa_0 + R_c^l \right) \mu_c^{\operatorname{T}} \Sigma_c^{-1} \mu_c + \left( \kappa_0 \mu_0 + \sum_n r_{nc}^l y^n \right)^{\operatorname{T}} \Sigma_c^{-1} \mu_c 
,\]
from which it is easy to see that $q^l(\mu_c \mid \Sigma_c) = \operatorname{N}(\mu_c \mid \mu_c^{l}, \kappa_{c}^{-l} \Sigma_c)$ according to the definitions of $\kappa_c^l$ and $\mu_c^l$. Also, we can now see that 
\begin{align*}
\log & \, q^l(\Sigma_c) \\
&= \log q^l(\mu_c, \Sigma_c) - \log q^l(\mu_c \mid \Sigma_c) \\
&\stackrel{\mathrm{c}}{=} -\frac{\nu_0 + \dim \mathcal{Y} + 2 + R_c^l}{2} \log \det \Sigma_c - \frac{1}{2} \operatorname{tr}(\Sigma_0 \Sigma_c^{-1}) \\
&\quad\quad - \frac{\kappa_0}{2} (\mu_c - \mu_0)^{\operatorname{T}} \Sigma_c^{-1} (\mu_c - \mu_0)  - \frac{1}{2} \sum_n r_{nc}^l (y^n - \mu_c)^{\operatorname{T}} \Sigma_c^{-1} (y^n - \mu_c) \\
&\quad\quad +\frac{1}{2} \log \det \Sigma_c + \frac{\kappa_c^l}{2} (\mu_c - \mu_c^l)^{\operatorname{T}} \Sigma_c^{-1} (\mu_c - \mu_c^l) \\
&= -\frac{\nu_0 + \dim \mathcal{Y} + 1 + R_c^l}{2} \log \det \Sigma_c \\
&\quad\quad - \frac{1}{2} \operatorname{tr}\!\Bigg( \Bigg( \Sigma_0 + \kappa_0 (\mu_c - \mu_0)(\mu_c - \mu_0)^{\operatorname{T}} + \sum_{n} r_{nc}^l (y^n - \mu_c) (y^n - \mu_c)^{\operatorname{T}} \\
&\quad\quad\quad\quad - \kappa_c^l (\mu_c - \mu_c^l)(\mu_c - \mu_c^l)^{\operatorname{T}} \Bigg) \, \Sigma_c^{-1} \Bigg) \\
&= -\frac{\nu_0 + \dim \mathcal{Y} + 1 + R_c^l}{2} \log \det \Sigma_c \\
&\quad\quad - \frac{1}{2} \operatorname{tr}\!\Bigg( \Bigg( \Sigma_0 + \kappa_0 \mu_0 \mu_0^{\operatorname{T}} + \sum_{n} r_{nc}^l y^n y^{n \mathrm{T}} - \kappa_c^l \mu_c^l \mu_c^{l \mathrm{T}} \Bigg) \, \Sigma_c^{-1} \Bigg)
\end{align*}
---recalling the definitions of $\nu_c^l$ and $\Sigma_c^l$, this amounts to $q^l(\Sigma_c) = \operatorname{IW}(\Sigma_c \mid \Sigma_c^l, \nu_c^l)$. Again, note that we have not made any assumptions about the forms of $q^l(\{(u^n, z^n)\}_n)$ and $q^l(\Lambda)$.

Regarding the positive definiteness of $\Sigma_c^l$ in relation to the $r_{nc}^l$, suppose first that $r_{nc}^l > 0$ for all $n$. Defining
\begin{align*}
K_c^l &= (\kappa_0 + R_c^l) \left( \kappa_0 \mu_0 \mu_0^{\mathrm{T}} + \sum_{n} r_{nc}^l y^n y^{n \mathrm{T}} \right) \\
&\quad\quad - \left( \kappa_0 \mu_0 + \sum_n r_{nc}^l y^n \right) \left( \kappa_0 \mu_0 + \sum_n r_{nc}^l y^n \right)^{\mathrm{T}},
\end{align*}
so that $\Sigma_c^l = \Sigma_0 + (\kappa_0 + R_c^l)^{-1} K_c^l$, we have for all vectors $\xi$ that
\begin{align*}
\xi^{\operatorname{T}} K_c^l \xi &= (\kappa_0 + R_c^l) \left( \kappa_0 (\xi^{\operatorname{T}} \mu_0)^2 + \sum_n r_{nc}^l (\xi^{\operatorname{T}} y^n)^2 \right) - \left( \kappa_0 \xi^{\operatorname{T}} \mu_0 + \sum_n r_{nc}^l \xi^{\operatorname{T}} y^n \right)^2 \\
&= \Vert 1 \Vert^2 \lVert \xi^{\operatorname{T}} \overline{y} \rVert^2 - \lvert \langle 1, \xi^{\operatorname{T}} \overline{y} \rangle \rvert^2 \\
&\geq 0
\end{align*}
by Cauchy--Schwarz---here, $\overline{y}$ is the row-augmented matrix $(\mu_0, (y^n)_n)$ and the inner product and norm are defined as $\langle x, x' \rangle = \kappa_0 x_0 x_0' + \sum_{i=1}^N r_{nc}^l x_i x_i'$ and $\Vert x \Vert^2 = \langle x, x \rangle$, which are well-defined as all $r_{nc}^l$ and $\kappa_0$ are positive. This shows the positive definiteness of $K_c^l$ and thus also $\Sigma_c^l$. 

\subsection{Proof of Proposition \ref{mstep2}}
\label{proofmstep2}

By the same procedure as in Appendices \ref{proofestep1}, \ref{proofmstep1}, the desired result follows from 
\begin{align*}
\log & \, q^{l+1}(\Lambda) \\
&\stackrel{\mathrm{c}}{=} \operatorname{\mathbb{E}}^{q^{l+1}(\{(\mu_c, \Sigma_c)\}_c) q^{l+1}(\{(u^n, z^n)\}_n)} \log p(\{(y^n, u^n, z^n)\}_n, \theta \mid \{x^n\}_n) \\
&\stackrel{\mathrm{c}}{\approx} \log \operatorname{W}(\Lambda \mid \Lambda_0, \eta_0) + \sum_n \sum_{n' \neq n} \sum_c \omega_{c, nn'}^{l+1} \Bigg( \log \operatorname{N}(x^n \mid x^{n'}, \Lambda^{-1}) \\
&\quad\quad - \sum_{n'' \neq n} t_{nn''}^{l+1} \log \operatorname{N}(x^n \mid x^{n''}, \Lambda^{-1}) \Bigg) \\
&\stackrel{\mathrm{c}}{=} \frac{\eta_0 - \dim \mathcal{X} - 1}{2} \log \det \Lambda - \frac{1}{2} \operatorname{tr}(\Lambda_0^{-1} \Lambda) + \sum_n \sum_{n' \neq n} \sum_c \omega_{c, nn'}^{l+1} \Bigg( \\
&\quad\quad + \frac{1}{2} \log \det \Lambda - \frac{1}{2}(x^n - x^{n'})^{\operatorname{T}} \Lambda (x^n - x^{n'}) \\
&\quad\quad - \sum_{n'' \neq n} t_{nn''}^{l+1} \left(\frac{1}{2} \log \det \Lambda - \frac{1}{2}(x^n - x^{n''})^{\operatorname{T}} \Lambda (x^n - x^{n''}) \right) \Bigg) \\
&= \frac{\eta_0 - \dim \mathcal{X} - 1}{2} \log\det \Lambda \\
&\quad\quad - \frac{1}{2}\operatorname{tr}\Bigg( \Bigg( \Lambda_0^{-1} + \sum_n \sum_{n' \neq n} (\Omega_{nn'}^{l+1} - t_{nn'}^{l+1})(x^n - x^{n'})(x^n - x^{n'})^{\operatorname{T}} \Bigg) \, \Lambda \Bigg),
\end{align*}
where we have used the facts that $\sum_{n' \neq n} \sum_c \omega_{c, nn'}^l = 1$ and $\sum_{n' \neq n} t_{nn'}^l = 1$.

\section{EM updates for local variational approximation}
\label{emememem}

Following Section \ref{optlinprog}, we are to maximize $\log \operatorname{\mathbb{E}}^{q^l(\theta)} p(\{(y^n, u^n, z^n)\}_n \mid \{x^n\}, \theta)$, where $p(\{(y^n, u^n, z^n)\}_n \mid \{x^n\}, \theta)$ is the linearized pseudolikelihood, with respect to the variational parameters $s$ corresponding to the second transitions. The form of the objective as an integral over the optimization parameters motivates the use of an EM algorithm, of which only one iteration is needed in Algorithm \ref{algo}.

Note that the rightmost expression in (\ref{lseapprox}) is invariant under uniform shifts of $\eta$, so we may assume without loss of generality that $\sum_i e^{\eta_i} = 1$. Letting $\zeta_i = e^{\eta_i}$ for all $i$, where $\zeta_i \geq 0$ and $\sum_i \zeta_i = 1$, we can rewrite (\ref{lseapprox}) as the equivalent approximation
\[
\operatorname{LSE}(\xi) \approx \sum_i \zeta_i (\xi_i - \log \zeta_i)
.\]
Using this, the first EM update at iteration $l$ in the mean-field variational Bayes algorithm may then be approximated as the minimizer of 
\begin{align*}
&\operatorname{\mathbb{E}}^{q^l(\theta)} \log p(\{(y^n, u^n, z^n)\}_n \mid \{x^n\}_n, \theta) \\
&\quad\quad \approx \operatorname{\mathbb{E}}^{q^l(\theta)} \sum_n \sum_{n' \neq n} \sum_c 1_{z^n \; = \; c} 1_{u^n \; = \; n'} \left( -\sum_{c'} s_{n'c'} \left( \log \operatorname{N}(y^{n'} \mid \mu_{c'}, \Sigma_{c'}) - \log s_{n'c'} \right) \right) \\
&\quad\quad = - \sum_n \left( \sum_{n' \neq n} \Omega_{n'n}^l \right) \sum_c s_{nc} \left( \operatorname{\mathbb{E}}^{q^l(\mu_c, \Sigma_c)} \log \operatorname{N}(y^{n} \mid \mu_{c}, \Sigma_{c}) - \log s_{nc} \right)
\end{align*}
over $s$ with nonnegative entries such that $\sum_{c} s_{nc} = 1$ for all $n$---that is, we should set the update $s^{l+1}$ to the $s$ for which each $s_n$ maximizes
\[
\sum_c s_{nc} \left( \operatorname{\mathbb{E}}^{q^l(\mu_c, \Sigma_c)} \log \operatorname{N}(y^{n} \mid \mu_{c}, \Sigma_{c}) - \log s_{nc} \right)
.\]
However, this amounts to solving $N$ separate constrained nonlinear optimization problems at each $l$, which would be too computationally expensive for most practical applications. We can circumvent this by removing the entropy term $-\sum_c s_{nc} \log s_{nc}$ above so that the optimization problem becomes linear---indeed, since the entropy term must be in the interval $[0, \log C]$ whereas the other linear term is unbounded, the expression may be regarded as approximately linear in $s_n$. Finally, in order to enforce the requirement of $r_{nc}^l = \sum_{n' \neq n} ( \omega_{c, nn'}^l + \omega_{c, n'n}^l - s_{nc}^l \Omega_{n'n}^l ) > 0$ in Proposition \ref{mstep1}, we add the constraint
\[
s_{nc} \leq \frac{\sum_{n' \neq n} (\omega_{c, nn'}^l + \omega_{c, n'n}^l)}{ \sum_{n' \neq n} \Omega_{n'n}^l}
\]
to each M-step in our EM algorithm and threshold the resulting $r_{nc}^l$ to be at least some small value. Thus, we recover the optimization problem (\ref{secondtransopt}) to optimize the second transition variational parameters. In particular, the fact that 
\begin{equation*}
\begin{split}
\operatorname{\mathbb{E}}^{q^l(\mu_c, \Sigma_c)} \log \operatorname{N}(y^n \mid \mu_c, \Sigma_c) &= \frac{1}{2} \sum_{i=1}^{\dim \mathcal{Y}} \psi\!\left( \frac{\nu_c^l + 1 - i}{2} \right) - \frac{1}{2} \log \det \Sigma_c^l  \\
&\quad\quad - \frac{1}{2} \frac{\dim \mathcal{Y}}{\kappa_c^l} - \frac{1}{2} \nu_c^l (y^n - \mu_c^l)^{\operatorname{T}} \Sigma_c^{-l} (y^n - \mu_c^l)
\end{split}
\end{equation*}
follows directly from Lemma \ref{thethree}.

\section{Evidence lower bound}
\label{appendixelbo}
We derive an explicit expression for $\operatorname{ELBO}(q^l)$, which may be evaluated at each iteration $l$ to monitor the progress of the variational approximation. Using the pseudolikelihood (\ref{pseudolikelihood}), we have the decomposition 
\begin{align*}
\operatorname{ELBO}(q^l) &= \operatorname{\mathbb{E}}^{q^l(\theta)} \log\!\left( \frac{p(\{(y^n, u^n, z^n)\}_n, \theta \mid \{x^n\}_n)}{q^{l}(\theta)} \right) \\
&\approx \sum_n \operatorname{\mathbb{E}}^{q^l(\theta)} \log p(y^n, u^n, z^n \mid \{(y^{n'}, u^{n'}, z^{n'})\}_{n' \neq n}, \{x^{n'}\}_{n'}, \theta) \\
&\quad\quad + \operatorname{\mathbb{E}}^{q^l(\theta)} \log p(\theta) - \operatorname{\mathbb{E}}^{q^l(\theta)} \log q^l(\theta).
\end{align*}
For the contribution from the full conditional, we again use the local variational approximations (\ref{secondtransapprox}) and (\ref{firsttransapprox}). With computations analogous to those in Appendices \ref{proofestep1}, \ref{proofmstep1} and \ref{proofmstep2}, we can write
\begin{align*}
\operatorname{\mathbb{E}}^{q^l(\theta)} &  \log p(y^n, u^n, z^n \mid \{(y^{n'}, u^{n'}, z^{n'})\}_{n' \neq n}, \{x^{n'}\}_{n'}, \theta) \\
&= -\frac{\operatorname{dim} \mathcal{Y}}{2} \log \pi - \frac{1}{2} \sum_c r_{nc}^l \Bigg( -\sum_{i=1}^{\operatorname{dim} \mathcal{Y}} \psi\!\left( \frac{\nu_c^l + 1 - i}{2} \right) + \log \det \Sigma_c^l \\
&\quad\quad + \frac{\operatorname{dim} \mathcal{Y}}{\kappa_c^l} + \nu_c^l (y^n - \mu_c^l)^{\operatorname{T}} \Sigma_c^{-l} (y^n - \mu_c^l) \Bigg) \\
&\quad\quad -\frac{1}{2} \eta^l \sum_{n' \neq n} (\Omega_{nn'}^l - t_{nn'}^l) (x^n - x^{n'})^{\operatorname{T}} \Lambda^l (x^n - x^{n'}).
\end{align*}
Furthermore, the contribution from the prior is computed as
\begin{align*}
&\operatorname{\mathbb{E}}^{q^l(\theta)}  \log p(\theta) \\
&\quad= \operatorname{\mathbb{E}}^{q^l(\Lambda)} \log \operatorname{W}(\Lambda \mid \Lambda_0, \eta_0) \\
&\quad\quad\quad + \sum_c \operatorname{\mathbb{E}}^{q^l(\mu_c, \Sigma_c)}\!\left[ \log \operatorname{N}\!\left( \mu_c \; \bigg| \; \mu_0, \frac{1}{\kappa_0} \Sigma_c \right) + \log \operatorname{IW}(\Sigma_c \mid \Sigma_0, \nu_0) \right] \\
&\quad= - \frac{\eta_0 \operatorname{dim} \mathcal{X}}{2} \log 2 - \frac{\eta_0}{2} \log \det \Lambda_0 - \log \Gamma_{\operatorname{dim} \mathcal{X}}\!\left( \frac{\eta_0}{2} \right) \\
&\quad\quad\quad + \operatorname{\mathbb{E}}^{q^l(\Lambda)}\!\left[ \frac{\eta_0 - \operatorname{dim} \mathcal{X} - 1}{2} \log \det \Lambda - \frac{1}{2}\operatorname{tr}(\Lambda_0^{-1} \Lambda) \right] \\
&\quad\quad\quad \sum_c \Bigg( -\frac{\operatorname{dim} \mathcal{Y}}{2} \log 2\pi \\
&\quad\quad\quad\quad\quad - \operatorname{\mathbb{E}}^{q^l(\mu_c, \Sigma_c)}\!\left[ -\frac{1}{2} \log \det\!\left( \frac{1}{\kappa_0} \Sigma_c \right) - \frac{\kappa_0}{2} (\mu_c - \mu_0)^{\operatorname{T}} \Lambda (\mu_c - \mu_0) \right] \\
&\quad\quad\quad\quad\quad -\frac{\nu_0 \operatorname{dim} \mathcal{Y}}{2} \log 2 + \frac{\nu_0}{2} \log \det \Sigma_0 - \log \Gamma_{\operatorname{dim} \mathcal{Y}}\!\left( \frac{\nu_0}{2} \right) \\
&\quad\quad\quad\quad\quad \operatorname{\mathbb{E}}^{q^l(\mu_c, \Sigma_c)}\!\left[ -\frac{\nu_0 + \operatorname{dim} \mathcal{Y} + 1}{2} \log \det \Sigma_c - \frac{1}{2} \operatorname{tr}(\Sigma_0 \Sigma_c^{-1}) \right] \Bigg) \\
&\quad= -\frac{\operatorname{dim} \mathcal{X} (\operatorname{dim} \mathcal{X} + 1)}{2} \log 2 + \frac{C \operatorname{dim} \mathcal{Y}}{2} ((\operatorname{dim} \mathcal{Y} + 1) \log 2 + \log \kappa_0 - \log \pi) \\
&\quad\quad\quad -\frac{\eta_0}{2} \log \det \Lambda_0 - \log \Gamma_{\operatorname{dim} \mathcal{X}}\!\left( \frac{\eta_0}{2} \right) + \frac{C \nu_0}{2} \log \det \Sigma_0 - C\log \Gamma_{\operatorname{dim} \mathcal{Y}}\!\left( \frac{\nu_0}{2} \right) \\
&\quad\quad\quad + \frac{\eta_0 - \operatorname{dim} \mathcal{X} - 1}{2}\left( \sum_{i=1}^{\operatorname{dim} \mathcal{X}} \psi\!\left( \frac{\eta^l + 1 - i}{2} \right) + \log \det \Lambda^l \right) - \frac{\eta^l}{2} \operatorname{tr}(\Lambda_0^{-1} \Lambda^l) \\
&\quad\quad\quad + \sum_c \Bigg( - \frac{\kappa_0}{2}\left( \frac{\operatorname{dim} \mathcal{Y}}{\kappa_c^l} + \nu_c^l (\mu_c^l - \mu_0)^{\operatorname{T}} \Sigma_c^{-l} (\mu_c^l - \mu_0) \right) \\
&\quad\quad\quad\quad\quad + \frac{\nu_0 + \operatorname{dim}\mathcal{Y} + 2}{2} \left( \sum_{i=1}^{\operatorname{dim} \mathcal{Y}} \psi\!\left( \frac{\nu_c^l + 1 - i}{2} \right) - \log \det \Sigma_c^l \right) \\
&\quad\quad\quad\quad\quad- \frac{\nu_c^l}{2} \operatorname{tr}(\Sigma_0 \Sigma_c^{-l}) \Bigg),
\end{align*}
and that from the variational posterior is computed analogously as
\begin{align*}
\operatorname{\mathbb{E}}^{q^l(\theta)} & \log q^l(\theta) \\
&= -\frac{\operatorname{dim} \mathcal{X} (\operatorname{dim} \mathcal{X} + 1)}{2} \log 2 + \frac{C \operatorname{dim} \mathcal{Y}}{2} ((\operatorname{dim} \mathcal{Y} + 1) \log 2 - \log \pi - 1) \\
&\quad\quad - \log \Gamma_{\operatorname{dim} \mathcal{X}}\!\left( \frac{\eta^l}{2} \right) - \frac{\eta^l \operatorname{dim} \mathcal{X}}{2} \\
&\quad\quad + \frac{\eta^l - \operatorname{dim} \mathcal{X} - 1}{2} \sum_{i=1}^{\operatorname{dim} \mathcal{X}} \psi\!\left( \frac{\eta^l + 1 - i}{2} \right) - \frac{\operatorname{dim} \mathcal{X} + 1}{2} \log \det \Lambda^l  \\
&\quad\quad + \sum_c \Bigg(  -\log \Gamma_{\operatorname{dim} \mathcal{Y}}\!\left( \frac{\nu_c^l}{2} \right) + \frac{ \operatorname{dim} \mathcal{Y}}{2} (\log \kappa_c^l - \nu_c^l) \\
&\quad\quad\quad\quad -\frac{\nu_c^l + \operatorname{dim} \mathcal{Y} + 2}{2} \sum_{i=1}^{\operatorname{dim} \mathcal{Y}} \psi\!\left( \frac{\nu_c^l + 1 - i}{2} \right) - \frac{\operatorname{dim}\mathcal{Y} + 2}{2} \log \det \Sigma_c^l  \Bigg).
\end{align*}

\section{Stochastic gate optimization}
\label{appedix_stochastic_gate_opt}

\subsection{Control variates for variance reduction}
\label{control_variates}

In Proposition \ref{stochgradgateopt} and in the gradient-based versions of Propositions \ref{estep} and the ELBO, expressions such as
\[
\operatorname{\mathbb{E}}^{p(A)} \operatorname{log}\!\left( \sum_{n' \neq n} \operatorname{exp}\!\left( -\frac{1}{2}(x^n - x^{n'})^{\operatorname{T}} L A A^{\operatorname{T}} L^{\operatorname{T}} (x^n - x^{n'}) \right) \right)
\]
reoccur. As the computation of the expression for each sample of $A$ is performance-critical, we devise control variates to reduce the variance compared to that of the standard Monte Carlo approximation. Since variance-minimizing weights cannot be obtained in closed forms, we will subtract a linearized version of the log-sum-exp expression and adjust with its analytically tractable expectation. In particular, note that 
\[
\operatorname{LSE}((\xi_i)_i) = \operatorname{log}\!\left( \sum_i e^{\xi_i} \right) \approx \operatorname{log}\!\left( \sum_i e^{\eta_i} \right) + \sum_i \frac{e^{\eta_i}}{\sum_{i'} e^{\eta_{i'}}} (\xi_i - \eta_i) \stackrel{\mathrm{c}}{=} \sum_i \zeta_i \xi_i
,\]
where $(\eta_i)_i$ is the location of the linearization and $(\zeta_i)_i = \operatorname{softmax}((\eta_i)_i)$. In our case, introducing $(\zeta_{nn'})_{n'\neq n}$ as the softmax-transformed location of the linearization for each $n$, the linearized log-sum-exp expression is equal up to a constant to 
\[
-\frac{1}{2} \sum_{n' \neq n} \zeta_{nn'} (x^n - x^{n'})^{\operatorname{T}} L A A^{\operatorname{T}} L^{\operatorname{T}} (x^n - x^{n'})
,\]
with expectation $-(\eta_0 / 2) \sum_{n' \neq n} \zeta_{nn'} (x^n - x^{n'})^{\operatorname{T}} L L^{\operatorname{T}} (x^n - x^{n'})$---thus, the corresponding estimator with control variates is written as
\begin{align*}
\operatorname{\mathbb{E}}^{p(A)}\!\Bigg[ & \operatorname{log}\!\Bigg( \sum_{n' \neq n} \operatorname{exp}\!\left( -\frac{1}{2}(x^n - x^{n'})^{\operatorname{T}} L A A^{\operatorname{T}} L^{\operatorname{T}} (x^n - x^{n'}) \right) \Bigg) \\
&\quad\quad\quad\quad + \frac{1}{2} \sum_{n' \neq n} \zeta_{nn'} (x^n - x^{n'})^{\operatorname{T}} L A A^{\operatorname{T}} L^{\operatorname{T}} (x^n - x^{n'}) \Bigg] \\
&- \frac{\eta_0}{2} \sum_{n' \neq n} \zeta_{nn'} (x^n - x^{n'})^{\operatorname{T}} L L^{\operatorname{T}} (x^n - x^{n'}).
\end{align*}
With $L^l$ being the Cholesky factor of the scale matrix at coordinate ascent iteration $l$, for the optimization of $L^{l+1}$, we may set the location of the linearization to the expected values of iteration $l$, leading to
\[
(\zeta_{nn'})_{n' \neq n} = \operatorname{softmax}\!\left( \left( -\frac{\eta_0}{2} (x^n - x^{n'})^{\operatorname{T}} L^l L^{l \mathrm{T}} (x^n - x^{n'}) \right)_{n' \neq n} \right)
.\]

\subsection{Center matrix computation}
\label{center_matrix}

In the optimization objective in Proposition \ref{stochgradgateopt}, the center matrix
\[
\Lambda_0^{-1} + \sum_{n} \sum_{n' \neq n} \Omega_{nn'}^l (x^n - x^{n'}) (x^n - x^{n'})^{\operatorname{T}}
\]
needs to be computed and stored in memory. To avoid evaluating the double sum of outer vector products by looping, it is easy to check that the following expression
\[
\Lambda_0^{-1} + x^{\operatorname{T}} \left( \operatorname{diag}(1^{\mathrm{T}} (\Omega^l + \Omega^{l \mathrm{T}})) - (\Omega^l + \Omega^{l \mathrm{T}}) \right) x
\]
produces the same matrix, where $x = (x^n)_{n}$ is the rowwise input data matrix and $\Omega^l = (\Omega^l_{nn'})_{n, n'}$. This allows for the use of more efficient vectorized matrix operations.

\section{Hyperparameter selection and variational parameter initialization}
\label{hyperparameter_selection}

Recall that the prior is written as
\[
p(\theta) = \operatorname{W}(\Lambda \mid \Lambda_0, \eta_0) \prod_c \operatorname{N}\!\left(\mu_c \; \bigg| \; \mu_0, \frac{1}{\kappa_0} \Sigma_c\right) \operatorname{IW}(\Sigma_c \mid \Sigma_0, \nu_0) 
,\]
where $\Lambda_0$, $\eta_0$, $\mu_0$, $\kappa_0$, $\Sigma_0$ and $\nu_0$ are hyperparameters. In our implementation, default choices for the hyperparameters are available, controlled by a handful of other, more intuitively hand-tuneable hyperparameters. In particular, the degree-of-freedom parameter $\eta_0 > \operatorname{dim} \mathcal{X} - 1$ for the Wishart-distributed $\Lambda$ is set to $\operatorname{dim} \mathcal{X} + \delta_{\operatorname{g}}$, where $\delta_{\operatorname{g}}$ is referred to as the excess degrees of freedom for the gate prior. In turn, the scale matrix $\Lambda_0$ is set to the inverse of the sample covariance of the input variable $x$, scaled by $\lambda_{\operatorname{g}} / \eta_0$ so that $\Lambda$ has expectation equal to the inverse of the sample covariance of $x$ up to a factor $\lambda_{\operatorname{g}}$. A larger $\delta_{\operatorname{g}}$ will imply a larger $\eta_0$, which will amplify the logarithmic determinant and trace terms in the objective in Proposition \ref{stochgradgateopt} and thus entail a higher degree of regularization of the \say{likelihood} term. Also, a higher $\lambda_{\operatorname{g}}$ will amplify $\Lambda_0^{-1}$ in relation to $\sum_n \sum_{n' \neq n} \Omega_{nn'}^l (x^n - x^{n'}) (x^n - x^{n'})^{\operatorname{T}}$ and, eventually, make the former dominate over the latter, in which case the scaling $\lambda_{\operatorname{g}}$ will roughly determine the magnitude of the center matrix. As $\operatorname{arg\,min}_A \; (-\operatorname{log} \operatorname{det} A + \operatorname{tr}(AB)) = B^{-1}$ for all positive definite $B$, the logarithmic determinant and trace terms will drive the scale matrix candidate $L L^{\operatorname{T}}$ toward the inverse of the center matrix. Thus, increasing $\lambda_0$ will in general have the effect of reducing the number of near neighbors of each input $x^n$. Similarly, for the experts, the degree-of-freedom parameter $\nu_0 > \operatorname{dim} \mathcal{Y} - 1$ is set to $\operatorname{dim} \mathcal{Y} + \delta_{\operatorname{e}}$, where $\delta_{\operatorname{e}}$ is the excess degrees of freedoms for the expert prior. The scale matrix $\Sigma_0$ is then set to $(\lambda_{\operatorname{e}} \nu_0) / C$ times the sample covariance of the output variable $y$, where $\lambda_{\operatorname{e}}$ is a scaling factor, so that each inverse-Wishart--distributed $\Sigma_c$ has expectation equal to the sample covariance up to a factor $\lambda_{\operatorname{e}} / C$. Moreover, the prior mean $\mu_0$ is set to the sample mean of $y$, and the related hyperparameter $\kappa_0 > 0$ is set to a small number to indicate a relatively weak belief that each $\mu_c$ will actually lie near $\mu_0$. 

Regarding the number $C$ of experts, as they are anchored at different locations across the output distribution, $C$ controls the fine-grainedness of the predictive distributions. Thus, in general, $C$ should be set as high as the computational resources allow, which is in the order $10^2$ for common computer setup---in the implementation, default is $C = 32$. If $C$ is on the large side of the actual model complexity, experts will conglomerate to produce a model which is effectively the same as one with a smaller $C$. Hence, reducing $C$ will reduce the set of possible predictive distributions, whereas an exaggerated $C$ will generally not affect the model outputs much as long as computational aspects are disregarded. An important point is that while the choice of $C$ is indeed a parameter to manually specify, thus moving away from a purely nonparametric paradigm, the situation is similar to that described in \citep{bleijordan}, where a finite mixture is used as a variational approximation to the infinite-mixture DPMM and the number of mixture classes needs to be explicitly specified. 

Apart from the hyperparameters of the priors, we also need to initialize the variational parameters for the mean-field variational Bayes iterations. At iteration $l = 0$, we set $\eta^0 = \eta_0$ and $\Lambda^0 = \Lambda_0$ for the gate. For the expert means, a clustering algorithm with a fixed number of $C$ clusters is used for assigning to each $\mu_c^0$ the mean in cluster $c$. For each expert $c$, we then set $\Sigma_c^0$, $\nu_c^0$ and $\kappa_c^0$ to the sample covariance of $y$, $\nu_0$ and a relatively high number, respectively, the latter to reflect the assertion that we do not want excessive movement in the expert means in order to avoid premature conglomeration of the mixture components. The variational parameters $s$ and $t$ are set such that $s_{nc} \propto 1$ and $t_{nn'} \propto 1$ with the required normalizations.

\end{appendices}

\printbibliography

\end{document}